\documentclass[runningheads]{llncs}
\usepackage{graphicx}

\usepackage{tikz}
\usepackage{comment}
\usepackage{color}
\usepackage{cite}
\usepackage{booktabs}
\usepackage{microtype}
\usepackage{makecell}
\usepackage{hyperref}
\usepackage{multirow}

\usepackage[accsupp]{axessibility}  

\newcommand\blfootnote[1]{%
  \begingroup
  \renewcommand\thefootnote{}\footnote{#1}%
  \addtocounter{footnote}{-1}%
  \endgroup
}

\begin{document}
\pagestyle{headings}
\mainmatter
\def\ECCVSubNumber{4715}  

\title{ParticleSfM: Exploiting Dense Point Trajectories for Localizing Moving Cameras in the Wild} 

\titlerunning{Exploiting Dense Point Trajectories for Localizing Moving Cameras}
%
\author{Wang Zhao\inst{1,3} \and
Shaohui Liu\inst{2,3} \and
Hengkai Guo\inst{3} \and \\
Wenping Wang\inst{4} \and
Yong-Jin Liu\inst{1}$^*$
}
\authorrunning{W. Zhao et al.}
%

\institute{$^1$Tsinghua University $^2$ETH Zurich $^3$ByteDance Inc.  $^4$Texas A\&M University}
\maketitle

\begin{abstract}
Estimating the pose of a moving camera from monocular video is a challenging problem, especially due to the presence of moving objects in dynamic environments, where the performance of existing camera pose estimation methods are susceptible to 
pixels that are not geometrically consistent. To tackle this challenge, we present a robust  \textit{dense indirect} structure-from-motion method for videos that 
is based on dense correspondence initialized from pairwise optical flow. Our key idea is to optimize long-range video correspondence as dense point trajectories and use it to learn robust estimation of motion segmentation. A novel neural network architecture is proposed for processing irregular point trajectory data. Camera poses are then estimated and optimized with global bundle adjustment over the portion of long-range point trajectories that are classified as static. 
Experiments on MPI Sintel dataset show that 
our system produces significantly more accurate camera trajectories compared to existing state-of-the-art methods. In addition, our method is able to retain reasonable accuracy of camera poses on fully static scenes, which consistently outperforms strong state-of-the-art dense correspondence based methods with end-to-end deep learning, demonstrating the potential of dense indirect methods based on optical flow and point trajectories. As the point trajectory representation is general, we further present results and comparisons on in-the-wild monocular videos with complex motion of dynamic objects. 
Code is available at \href{https://github.com/bytedance/particle-sfm}{\color{black}{https://github.com/bytedance/particle-sfm}}.

\blfootnote{$^*$Corresponding author.}

\end{abstract}

\section{Introduction}

\begin{figure}
    \centering
    \includegraphics[trim={0 100 0 40}, clip,  width=0.99\linewidth, height=120pt]{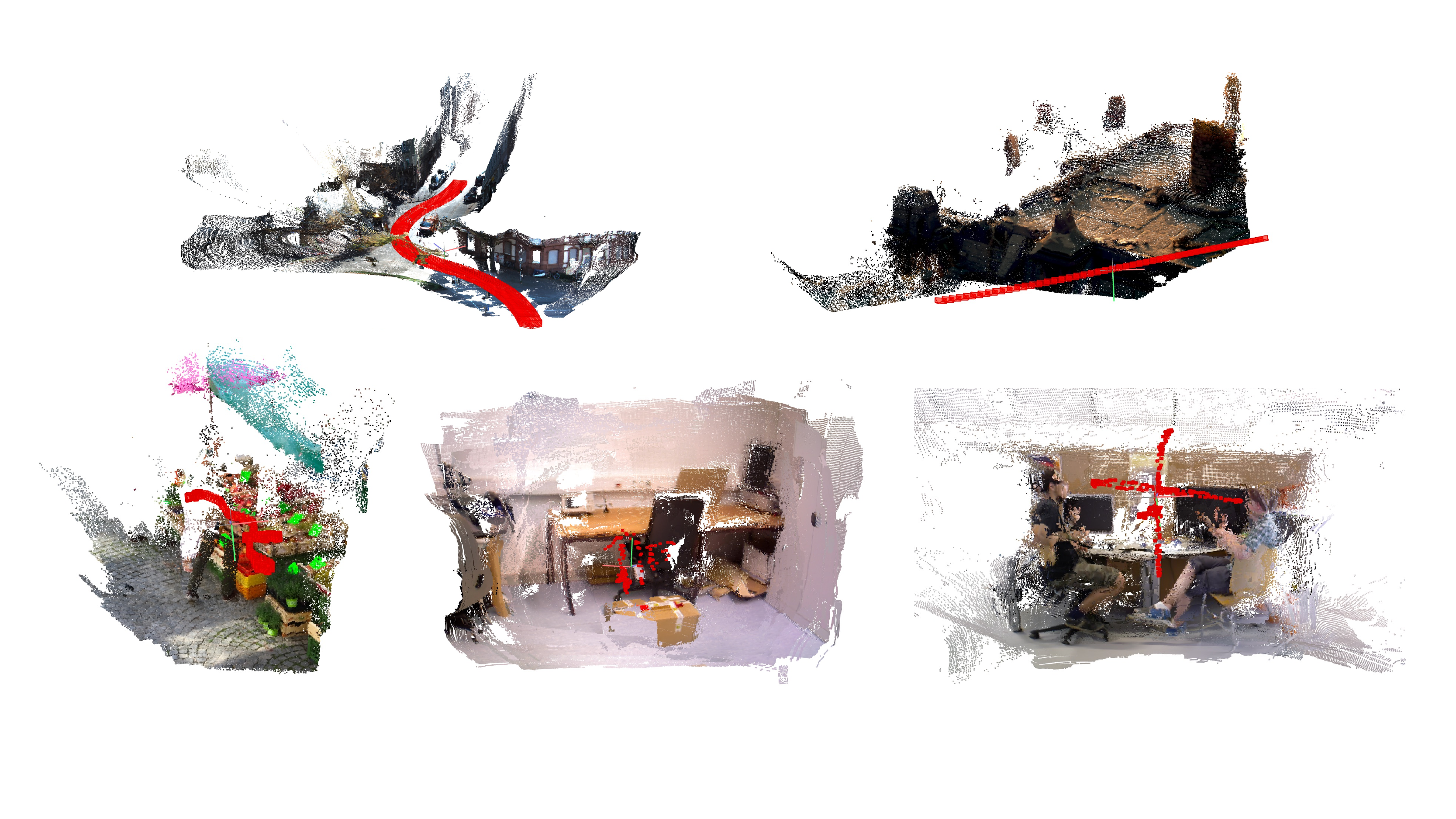}
    \caption{We present a dense indirect method that is able to recover reliable camera trajectories from in-the-wild videos with complex object motion in dynamic scenes. (See Fig. \ref{fig::wild} for qualitative comparisons with COLMAP \cite{schonberger2016structure})  }
    \label{fig::teaser}
\end{figure}

Localizing moving cameras from monocular videos 
is a fundamental task in a variety of applications such as augmented reality and robotics. Many videos exhibit complex foreground motion from humans, vehicles and general moving objects, posing severe challenges for robust camera pose estimation. Traditional indirect SfM methods \cite{klein2007parallel,forster2014svo,mur2015orb} are built on top of feature point detectors and descriptors. These methods rely on high-quality local features and perform non-linear optimization over the geometric reprojection error. Conversely, direct methods \cite{newcombe2011dtam,engel2014lsd,engel2017direct} track cameras by optimizing photometric error of the full image assuming consistent appearance across views. While both types of methods have produced compelling results, neither is robust against large object motion in dynamic environments, which is however ubiquitous in daily videos.

To mitigate the influence of moving objects, several existing monocular SLAM and SfM methods ~\cite{yu2018ds, yang2019cubeslam, bescos2018dynaslam} attempt to focus on specific semantic classes of objects that are likely to move around, e.g. humans and cars. However, there are many general objects that can possibly move in the scene in practice (e.g. a chair carried by a human). And moreover, those ``special'' objects such as humans and cars are not necessarily moving in the videos, making these semantics-based methods limited. Recent methods employ end-to-end deep learning to implicitly deal with those complex motion patterns, placing focus on static parts with the aid of training data. However, the end-to-end learning on camera poses brings up limitation on the system generalization on in-the-wild daily videos. 

We present a new \textit{dense indirect} structure-from-motion system for videos that explicitly tackles the issues brought by general moving objects. Our method is based on dense correspondence initialized from pairwise optical flow. Inspired by the success of Particle video \cite{sand2008particle}, our method exploits long-range video correspondence as dense point trajectories, which serves as an intermediate representation and provides abundant information for estimating motion segmentation and optimizing cross-view geometric consistency at global bundle adjustment. 

Specifically, our method first connects and optimizes dense point trajectories using pairwise dense correspondence from optical flow. Then, we propose a specially designed network architecture to learn robust estimation of motion labels from point trajectories with variable lengths. 
Finally, we apply global bundle adjustment to estimate and optimize camera poses and maps over portions of each point trajectory that are classified as static. 
Since the point trajectories are high-level abstraction of the input monocular videos, training motion estimation solely on synthetic datasets such as FlyingThings3D~\cite{mayer2016large} exhibits great generalization ability and enables our system to produce robust camera trajectories on general videos that contain complex and dense motion patterns (as shown in Fig. \ref{fig::teaser}).

Experiments on MPI Sintel dataset~\cite{butler2012naturalistic} validate that our method significantly improves over state-of-the-art SfM pipelines on localizing moving cameras in dynamic scenes. In addition, on full static ScanNet dataset~\cite{dai2017scannet}, our method is able to retain reasonable accuracy on the predicted camera trajectories, consistently outperforming strong dense correspondence based methods with end-to-end deep learning, such as DROID-SLAM \cite{teed2021droid}. We further present results and comparisons on in-the-wild monocular videos to demonstrate the improved robustness on dealing with complex object motion in dynamic environments.

\section{Related Work}
\textbf{Dense Correspondence in Videos: }
Cross-image correspondences are conventionally built on local feature point detection and description \cite{lowe2004distinctive,rublee2011orb}, and recent methods employ deep neural networks to improve local features \cite{detone2018superpoint,dusmanu2019d2} and matching \cite{sarlin2020superglue}. While the de-facto methods for localization and mapping operate on sparse feature points, dense pixel correspondences \cite{rocco2018neighbourhood,germain2020s2dnet,sun2021loftr} have shown great potential especially on videos, thanks to the rapid developments of optical flow predictors. Early methods such as SIFT-Flow~\cite{liu2010sift} achieve dense correspondences across different scenes, while recent advances~\cite{dosovitskiy2015flownet, ilg2017flownet, sun2018pwc} with deep learning regress the optical flow through differentiable warping and feature cost volumes. 
One notable recent work is RAFT~\cite{teed2020raft} that introduces iterative recurrent refinement with strong cross-dataset generalization. 
While these optical flow methods have achieved remarkable performance on per-pixel accuracy, they intrinsically limit the correspondences to image pairs rather than the whole video data. Conversely, long-range video correspondence is early studied in Particle video~\cite{sand2008particle} which first employs point trajectories to represent motion patterns in videos. However, the method is very computational expensive and practically not suitable for long videos. Sundaram et al.~\cite{sundaram2010dense} further propose a fast parallel implementation of variational large displacement optical flow~\cite{brox2010large}, and directly accumulate optical flow to get dense point trajectories. Inspired by these pioneer works, we present a method that sequentially tracks the optical flow and optimizes the pixel locations by exploiting path consistency to acquire reliable dense point trajectories.

~\\
\noindent
\textbf{Motion Segmentation:}
Motion segmentation aims to predict what is move for each image in a video sequence. Classical methods~\cite{shi1998motion, wang1994representing} estimate the motion mask based on optical flow analysis, with follow-up works~\cite{brox2006variational, cremers2005motion} formulating joint optimization over optical flow and motion segmentation. Recent methods with deep neural networks ~\cite{tokmakov2019learning, jain2017fusionseg, zhou2020motion} extract both appearance and optical flow features with a specially designed two-branch networks, or exploit semantic information~\cite{dave2019towards} with optical flow based motion grouping.
Self-supervised methods~\cite{yang2019unsupervised, yang2021self} are also proposed to overcome the requirement of labeling data. To better parse the object motion, relative camera motion is estimated and incorporated in~\cite{bideau2016s, bideau2018moa, lamdouar2020betrayed, yang2021learning}. 
However, these optical-flow based methods mostly suffer from temporal inconsistency and in-the-wild generalization to dynamic videos with different shapes and textures.
On the other hand, point trajectories contain rich information for temporal object motion. There are a number of existing works~\cite{sheikh2009background, lezama2011track, fragkiadaki2012video, ochs2012higher, ochs2013segmentation, keuper2015motion} that cluster the tracked point trajectories into motion segments with hand-crafted features and measurements. However, all theses clustering-based methods rely on heavy optimization and hand-crafted design, making them less scalable and general on in-the-wild video sequences. In this work, we combine the best of both worlds and introduce a specially designed network architecture for predicting trajectory attributes, specifically motion labels from irregular data of dense point trajectories. 

~\\
\noindent
\textbf{SfM and SLAM:}
Traditional structure-from-motion methods can be generally classified into indirect and direct methods. Indirect approaches \cite{klein2007parallel,forster2014svo,mur2015orb, schonberger2016structure, resch2015scalable, wilson2014robust, moulon2013global, yokozuka2019vitamin} rely on matched salient keypoints to determine geometric relationship for multi-view images. Conversely, direct methods \cite{newcombe2011dtam,engel2014lsd,engel2017direct,yang2017direct,alismail2016photometric} approximate gradients over dense photometric registration on the full image. While both trends of methods achieve great success in practice, they both suffer in dynamic scenes due to the large number of pixel outliers. Most related to us, recent literature attempts to exploits dense correspondences from optical flow. In particular, \cite{ranjan2019competitive,zhao2020towards} employ two-view geometric consistency check to detect moving objects, and visual odometry systems are built by triangulating optical flow correspondences in \cite{zhao2020towards,zhan2020visual}. VOLDOR \cite{min2020voldor} employs a probabilistic graphical model over optical flow to recover camera poses as hidden states. In TartanVO \cite{wang2020tartanvo}, optical flow is fed into an end-to-end network to directly predict camera poses. R-CVD~\cite{kopf2021robust} jointly optimizes depth and pose with flexible deformations and geometry-aware filtering. DROID-SLAM~\cite{teed2021droid} implicitly learns to exclude dynamic objects by training on sythetic data with dynamic objects and employing deep bundle adjustment over keyframes. While our method also benefits from dense correspondences from pairwise optical flow, we do not employ end-to-end deep learning to encourage better generalization. Instead, we explicitly model dense point trajectories and perform bundle adjustment over these video correspondences.

~\\
\noindent
\textbf{Localizing Cameras in Dynamic Scenes:}
Localizing cameras in dynamic scenes is challenging due to violation of rigid scene assumption in multi-view geometry. Classical SfM and SLAM methods reject moving pixels as outliers through robust cost function~\cite{klein2007parallel} and RANSAC~\cite{mur2015orb, schonberger2016structure}, yet consistently fail under highly dynamic scenes with complex motion patterns. Beyond monocular SLAM, effective segmentation and tracking of dynamic objects can be achieved~\cite{alcantarilla2012combining, wang2014motion, kim2016effective, sun2017improving, li2017rgb, barsan2018robust, huang2020clustervo} with auxiliary depth data from stereo, RGB-D and LiDAR, which, however, is not generally available for in-the-wild captured videos. Thanks to the rapid development of deep learning on visual recognition, many works~\cite{yu2018ds, yang2019cubeslam, bescos2018dynaslam, zhang2020vdo, ballester2021dot} tackle this problem by exploring the combination with object detection, semantic and instance segmentation. However, These methods are often restricted to pre-defined semantic classes. On the contrary, our method exploits the potential of long-range point trajectories in videos for robust motion label estimation. 

\section{Methods}
\begin{figure}[t]
    \centering
    \includegraphics[trim={0 30 0 20}, clip, width=0.99\linewidth, height=120pt]{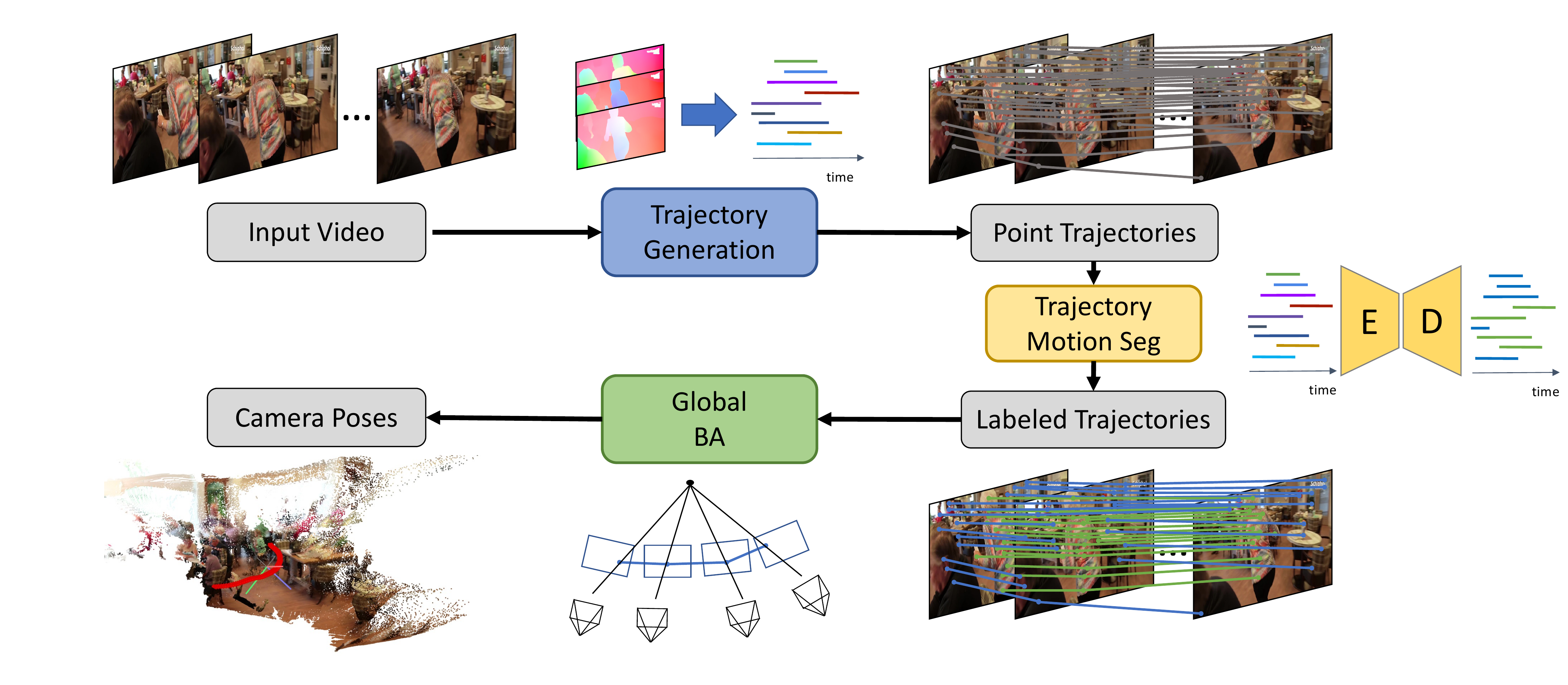}
    \caption{Overview of our proposed system for localizing moving cameras. Given an input video, we first accumulate and optimize over pairwise optical flow to acquire high-quality dense point trajectories. Then, a specially designed network architecture is employed to process irregular point trajectory data to predict per-trajectory motion labels. Finally, the optimized dense point trajectories along with the motion labels are exploited for global bundle adjustment (BA) to optimize the final camera poses}
    \label{fig::pipeline}
\end{figure}

The core idea of the proposed system is to exploit long-range video correspondences as dense point trajectories throughout the pipeline. Figure \ref{fig::pipeline} shows an overview of the system. We first accumulate and optimize point trajectories sequentially in a sliding-window manner from optical flow. Then, those point trajectories are fed into the specially designed trajectory processing network to predict per-trajectory motion labels. 
Finally, we estimate initial camera poses, triangulate global maps from the portions of each trajectory that are classified as static, and perform bundle adjustment over those point tracks to optimize both camera poses and 3d points.
As our method employs dense correspondence, the maps built from our system are denser and more complete than top sparse indirect methods such as COLMAP \cite{schonberger2016structure} with comparable running time needed.

\subsection{Acquiring Dense Point Trajectories}

We aim to acquire reliable point trajectories for motion estimation and global bundle adjustment. Following the practice of early literature~\cite{sand2008particle, sundaram2010dense} that focus on long-range video correspondence, we start from pairwise optical flow and sequentially accumulate them into point trajectories. We use RAFT \cite{teed2020raft} as the base optical flow predictor. 

For accumulation of dense point trajectories, given the current pixel location $p_0$ on image 0 (sized $H \times W$) and the optical flow $F_{0\rightarrow 1} \in \mathbb{R}^{H\times W\times 2}$ from image 0 to image 1, the trajectory can be extended with: $p_1 = p_0 + F_{0\rightarrow 1}(p_0)$. We continue tracking point trajectories until the point suffers from occlusion, which is determined by the forward-backward optical flow consistency check following common practice~\cite{sundaram2010dense, yin2018geonet}. This forward-backward consistency check not only deals with occlusion, but also filters out some erroneous optical flow, making the trajectories more reliable. To maintain dense point trajectories, for each accumulation step we generate new trajectories on the area that is not occupied by any trajectory on the current image. All trajectories are initialized at grid points. Following \cite{sundaram2010dense}, a sub-sampling factor $\lambda$ is employed to control the density of the trajectory by sub-sampling unoccupied pixels uniformly on 2D space, which helps balance the computational cost and trajectory density if needed.

The resulting point trajectories are dense and roughly correct as video correspondences. However, small errors from the optical flow accumulates across time, incurring the tracked pixel locations to gradually drift away from the true ones. Similar drifting errors often occur for tracking-based methods~\cite{zhang2015visual, kim2018low} for accumulating sensor measurements. To fight against drifting, Particle video~\cite{sand2008particle} attempts to perform a heavy optimization on pixel locations directly using appearance error on raw intensity images. Different from theirs, our method exploits path consistency that benefits from optical flow from non-adjacent image pairs.

Specifically, given consecutive frames ${I_0, I_1, I_2}$ and pairwise optical flows ${F_{i\rightarrow j}}$ from $I_i$ to $I_j$, we first initialize the point trajectory $p_1, p_2$ on $I_1$ and $I_2$ sequentially with direct accumulation:
\begin{equation}
    p_1' = p_0 + F_{0\rightarrow 1}(p_0), \ \  
    p_2' = p_1' + F_{1\rightarrow 2}(p_1').
\end{equation}

Then, we compute stride-2 optical flow $F_{0\rightarrow 2}$ and optimizes $p_1$ and $p_2$ with respect to the following objectives:
\begin{equation}
    \begin{aligned}
    L = (p_1 - p_1')^2 + & (p_2 - (p_0 + F_{0\rightarrow 2}(p_0)))^2 \\
    & + (p_2 - (p_1 + F_{1\rightarrow 2}(p_1)))^2
    \end{aligned}
\end{equation}
Gradients are numerically tractable with interpolation on the optical flow map $F_{1\rightarrow 2}$. Through the optimization we jointly adjust the pixel locations along the track to encourage consistency. The framework is easily extended for longer windows to exploit longer range of optical flow correspondences but we empirically find that adding a single stride-2 constraints already consistently improves the track quality by mitigating the drifting problem.

It is also worth noting that the success of path consistency formulation relies on the key assumption that the direct pairwise measurements are relatively more accurate than cumulative results when constructing point trajectories. This is generally true for accumulating sensor measurements for visual odometry, while in our case the assumption can be violated because if the pixel motion between two images is too significant the accuracy of the pairwise optical flow will degrade. Thus, it is relatively safe to keep a small window to avoid large degradation of the long-range optical flow. In practice, if $F_{0 \rightarrow 2}$ does not pass forward-backward consistency check, we skip the optimization at the timestamp and keep the initial accumulated positions. The point trajectory is sequentially extended and optimized until occlusion is detected.

\subsection{Trajectory-based Motion Segmentation}
\begin{figure}[t]
    \centering
    \includegraphics[trim={0 40 0 30}, clip, width=0.99\linewidth]{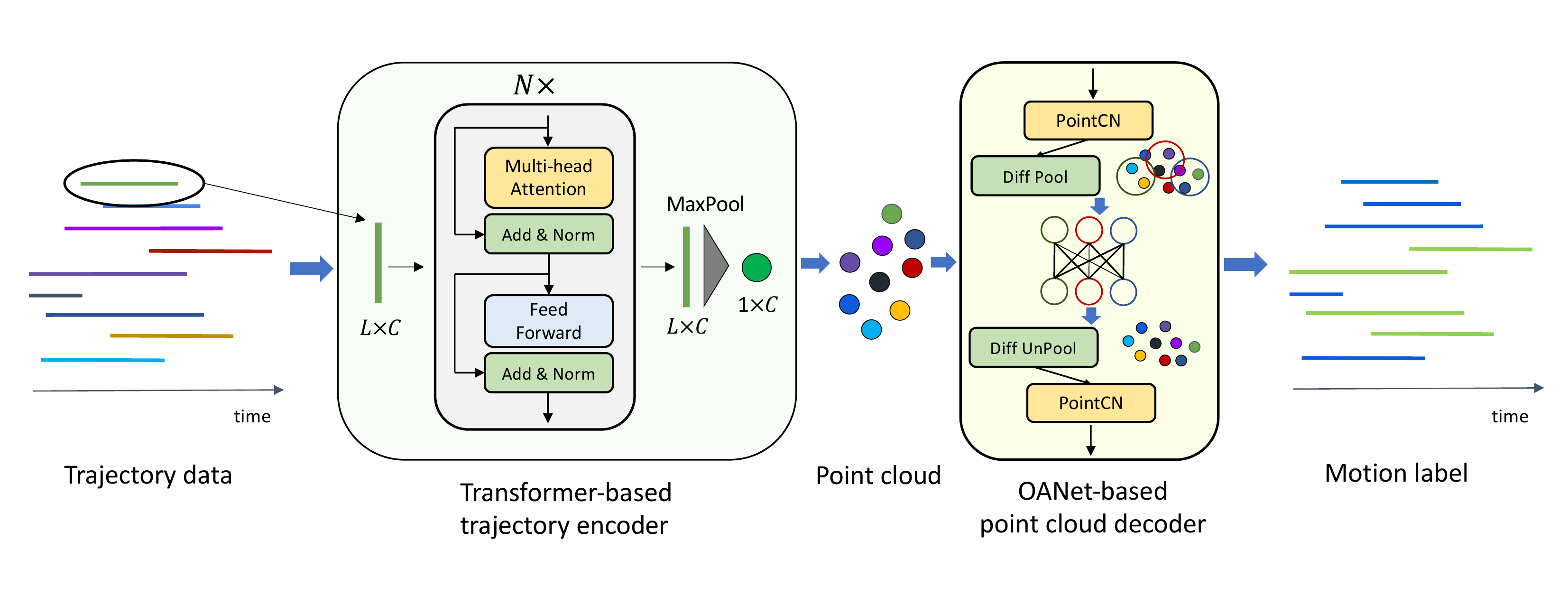}
    \caption{Illustration of the trajectory-based motion segmentation network. To process irregular point trajectory data, we first employ a transformer-based encoder to extract features for each trajectory. Then, all trajectory features are considered as high-dimensional point cloud and fed into an OANet-based \cite{zhang2019learning} decoder to predict per-trajectory motion labels. In practice, we split long-range point trajectories into segments and map the predicted motion labels for each segment back onto the pixels}
    \label{fig::network}
\end{figure}

Moving objects, while being ubiquitous in daily videos, pose severe challenges for camera pose estimation as it violates geometric consistency across different timestamps, making it crucial to design strategies to filter out dynamic objects. One trend of commonly-used methods in dynamic SLAMs~\cite{yu2018ds, yang2019cubeslam, bescos2018dynaslam} is to utilize semantic segmentation model, e.g. Mask-RCNN~\cite{he2017mask} to get per-pixel semantic masks, and remove all potentially moving pixels according to its semantic labels, such as person, car, dog, etc. While these methods are effective under certain scenarios, they are intrinsically not general for segmenting moving objects, since it completely relies on pre-defined semantic classes and cannot distinguish moving or static objects within the same semantic category (e.g. moving and static cars). Conversely, another alternative is to use two-frame motion segmentation methods~\cite{tokmakov2019learning, zhou2020motion}, which are recently powered with various convolutional neural network (CNN) models. These methods provide true motion labels instead of semantic candidates. However, two-frame based CNN models suffer from severe temporal inconsistency, degraded estimation when input flows are noisy, and exhibit limited out-of-domain generalization on in-the-wild videos.

We propose to exploit the dense point trajectories we acquired for estimating motion labels. Trajectory-based methods are generally more robust and consistent compared to two-frame flow models for motion segmentation. Unfortunately, traditional cluster-based trajectory segmentation methods rely on heavy optimization and hand-crafted features, and are hard to scale with dense trajectories. To deal with such issues, we propose a novel neural network for processing trajectory data for prediction, which enables fast, robust and accurate dense point trajectory based motion segmentation. As shown in Fig. \ref{fig::network}, our proposed network employs a encoder-decoder architecture. Specifically, the encoder directly consumes irregular trajectory data and embeds into high-dimensional feature space. Then, all the encoded trajectories are together fed into the decoder, which performs context-aware feature aggregations among trajectories to fuse both local and global information and finally regress the motion label. 

As each trajectory has different start time, end time and length, the point trajectory data is highly irregular and hard to be directly processed with regular convolutional neural networks. Inspired by sequence modeling in natural language processing where language sequences are also irregular, we utilize the powerful transformer~\cite{vaswani2017attention} model to extract features from trajectories. Built up with multi-head attention, transformer can effectively process sequential data, and is broadly used in language model, and recently extended to vision tasks~\cite{dosovitskiy2020image}. Our input data for transformer encoder is $N$ trajectories, and each trajectory includes a set of normalized pixel coordinates ${(u_i,v_i)}$. We first cut and pad all the trajectories to the temporal window size $L$. After that, all trajectories have the shape of $(L,2)$ with masks indicating where the pixel coordinates are zero-padded. To better exploit motion information, we augment the trajectory data $\{(u_i, v_i), i\in[0,L)\}$
with consecutive motion $(\Delta u_i, \Delta v_i) = (u_{i+1} - u_i, v_{i+1} - v_i)$. Furthermore, since the moving objects are much easier to be classified in 3d space, we integrate relative depth information from MiDaS \cite{ranftl2019towards} to disambiguate the motion segmentation from pure 2d pixel movements. The estimated relative depth is normalized to $(0,1)$ and used to back-project the 2d pixels into 3d camera coordinates $(x_i, y_i, z_i)$. We also include the 3d motion data $(\Delta x_i, \Delta y_i, \Delta z_i) = (x_{i+1} - x_i, y_{i+1} - y_i, z_{i+1} - z_i)$ for the trajectory input. The final augmented trajectories have the shape of $(L, 10)$. This data is then embedded by two MLPs, resulting in intermediate features with shape $(L,C)$, which is fed into the transformer module. The transformer consists of 4 blocks, each with multi-head attention and feed-forward layers to encode the temporal information of each trajectory. The output of the transformer module is encoded features also with shape $(L,C)$. To get feature representation of the whole trajectory instead of each point, we further perform max-pooling over temporal dimension, resulting in $(1,C)$ feature vector for each trajectory.

To segment a trajectory as moving objects or static background, the model not only needs to extract motion pattern from each trajectory data, but also has to communicate and compare with other trajectories before making decisions, which is achieved by the decoder. All encoded trajectory features with shape $(N, C)$ naturally forms a feature cloud in high-dimensional space. Thus, we build the decoder on top of carefully designed point cloud processing architectures. 

Specifically, we choose OANet~\cite{zhang2019learning} as our backbone to benefit from its efficiency and well-designed local-global context mechanism. To capture the local feature context, the network first clusters input points by learning a soft assignment matrix (Diff Pool in Figure~\ref{fig::network}).  Then, the clusters are spatially correlated to explore the global feature context. Next, the detailed context features of each point are recovered from embedded clusters through differentiable unpooling. And finally, several PointCN layers are applied, followed by sigmoid activation to get the binary prediction mask. We refer the reader to ~\cite{zhang2019learning} for more details about differentiable pooling, unpooling and PointCN layer.

Our proposed trajectory motion segmentation network is fast, robust and general. It could process tens of thousands trajectories within a few seconds. By only trained on synthetic dataset \textit{FlyingThings3D}~\cite{mayer2016large}, the network generalizes well across various scenarios including indoor, outdoor, synthetic movies and daily videos. Furthermore, the proposed network can also be easily extended to predict other trajectory attributes that benefit video understanding.

\subsection{Global Bundle Adjustment over Point Trajectories}
The optimized dense point trajectories along with the predicted motion labels can be jointly used for localizing moving cameras for monocular videos. Specifically, given a video sequence, we first accumulate and optimize optical flow to get long-range point trajectories. Then, the trajectory motion segmentation network predicts trajectory motion labels in a sliding-window manner with window size $L$.
These motion labels are mapped back onto each pixel to get final per-point motion segmentation. In the map construction and global bundle adjustment we only consider the portions of each trajectory that have static labels.

Since we have dense video correspondences, we can directly formulate non-linear geometric optimization over the point tracks. Inspired by \cite{wilson2014robust,theia-manual}, we build a global SfM pipeline with dense point trajectories. Specifically, relative camera poses for neighboring views are first solved \cite{hartley1997defense} with sampled correspondences from pairs of static pixels from the point trajectories with its motion labels. Then, rotation averaging \cite{chatterjee2013efficient} and translation averaging \cite{ozyesil2015robust} are performed to get the initial camera pose estimations. Finally, global bundle adjustment is applied over the constructed point tracks at triangulation stage. Note that the point tracks are consistent with the original dense point trajectories since it comes from dense correspondences sampled from it. Thus, we achieve final pose refinement by making use of the static pixels along the trajectories without considering outlier pixels that are classified as parts of moving objects. As the trajectory processing network is well generalized, our method can recover reliable camera trajectories on in-the-wild daily videos with complex foreground motion.

\section{Experiments}
\subsection{Implementation Details}
We use the pre-trained RAFT~\cite{teed2020raft} model on \textit{FlyingThings3D} dataset~\cite{brox2010large}. The sub-sampling factor $\lambda$ of point trajectory is set to 2 for all the experiments to balance the reconstruction density and computation time. 
For trajectory motion segmentation network, we use 4 heads for multi-head attention and 64 dimension for feed-forward layer. As for OANet, we set the cluster number to 100 and the layer number to 8. We train the proposed network on the training split of \textit{FlyingThings3D} dataset~\cite{mayer2016large}. We implement the network in PyTorch~\cite{paszke2017automatic} and train it using Adam optimizer~\cite{kingma2014adam} with learning rate 1e-4 for 30 epochs. At inference, we use a window size $L$ of 10 by default. 

We test our system on Sintel~\cite{butler2012naturalistic}, ScanNet~\cite{dai2017scannet}, and in the wild video sequences from DAVIS~\cite{perazzi2016benchmark}. For ScanNet dataset, we evaluate our method on the first 20 scenes in the test split. Since the whole video sequence is very long and not suitable for evaluating time-consuming offline methods like COLMAP~\cite{schonberger2016structure}, we only take the first 1500 frames and down-sample with stride 3 for each scene, resulting in a roughly 10 FPS video. 
Since all the methods use monocular video as input, we first scale and align all the output camera trajectories with respect to groundtruth, and then calculate commonly used pose metrics: RMSE of absolute trajectory error, translation and rotation part of relative pose error.

\begin{figure*}[tb]
\scriptsize
\centering
\begin{tabular}{ccccc}
\centering
{\includegraphics[width=0.17\linewidth]{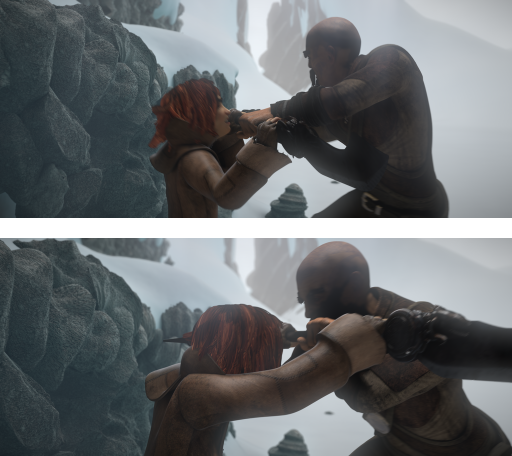}} & 
{\includegraphics[width=0.17\linewidth]{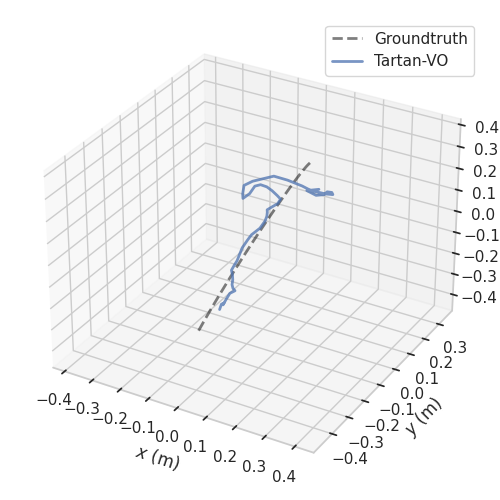}} & 
{\includegraphics[width=0.17\linewidth]{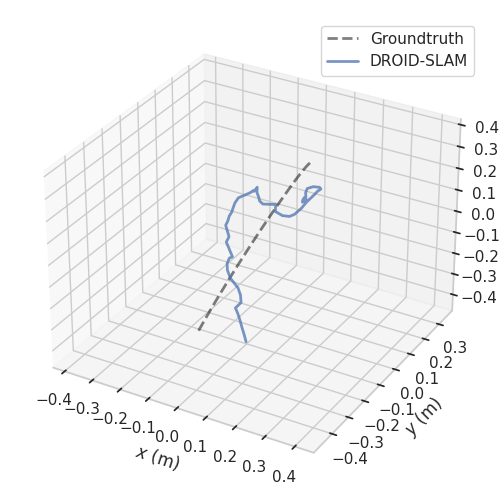}} & 
{\includegraphics[width=0.17\linewidth]{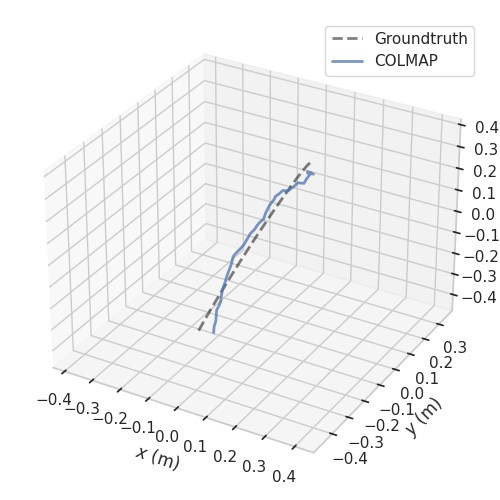}} &
{\includegraphics[width=0.17\linewidth]{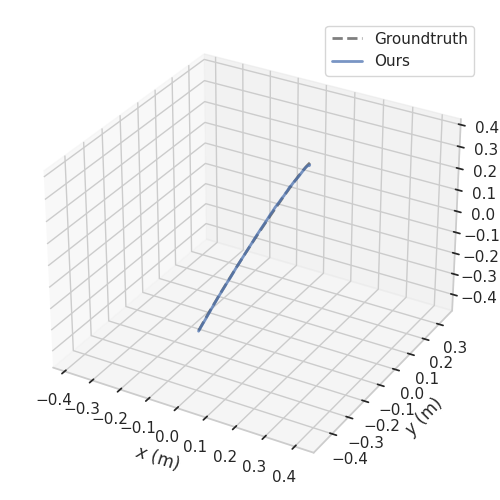}} \\

{\includegraphics[width=0.17\linewidth]{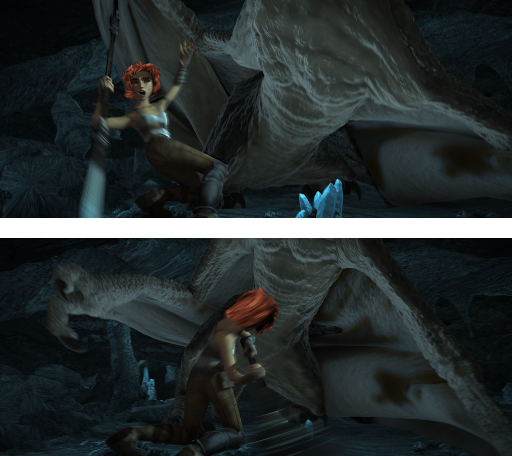}} & 
{\includegraphics[width=0.17\linewidth]{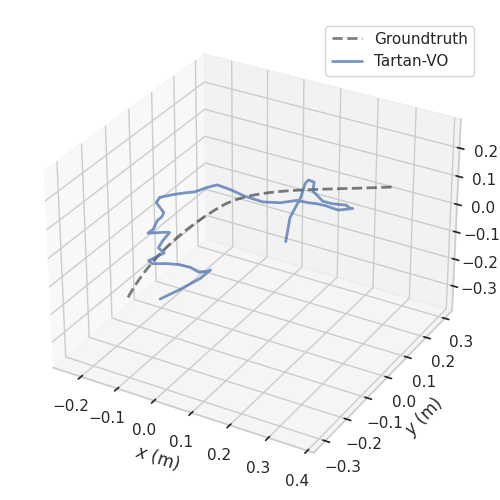}} & 
{\includegraphics[width=0.17\linewidth]{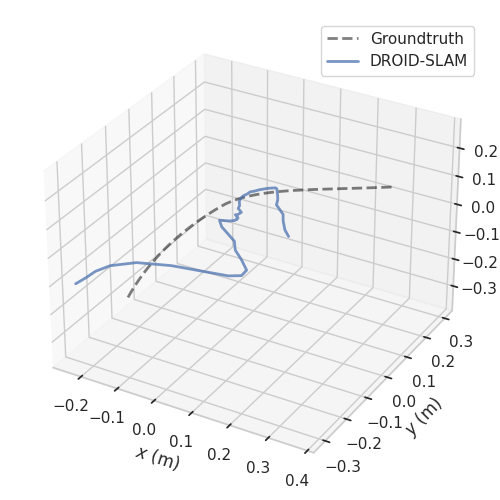}} & 
{\includegraphics[width=0.17\linewidth]{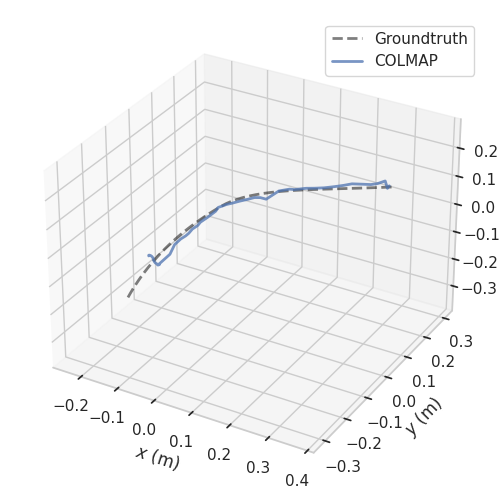}} &
{\includegraphics[width=0.17\linewidth]{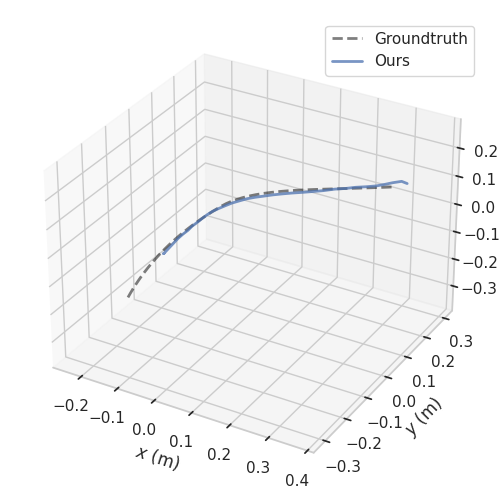}} \\

{\includegraphics[width=0.17\linewidth]{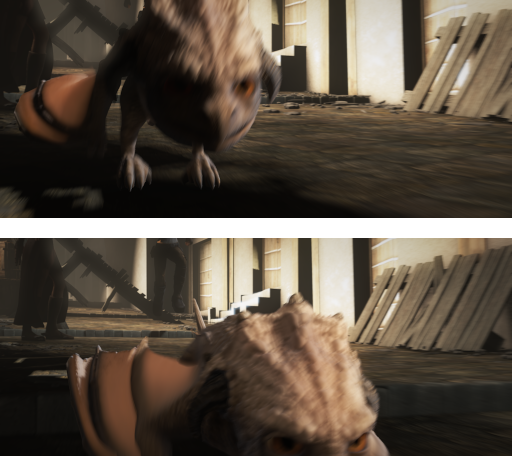}} & 
{\includegraphics[width=0.17\linewidth]{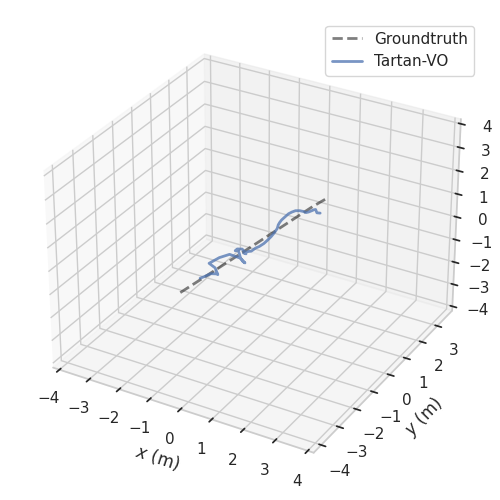}} & 
{\includegraphics[width=0.17\linewidth]{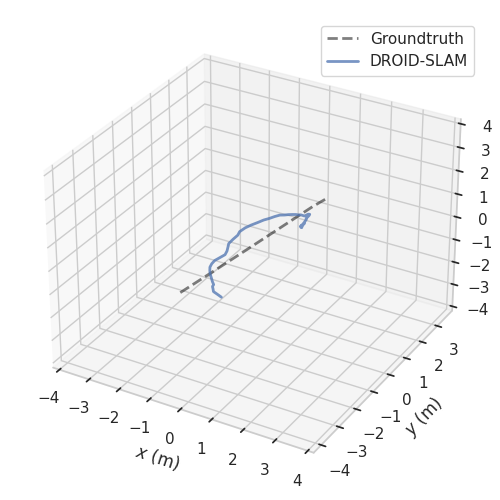}} & 
{\includegraphics[width=0.17\linewidth]{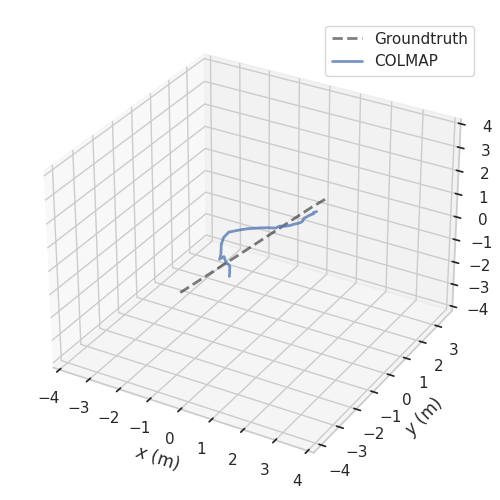}} &
{\includegraphics[width=0.17\linewidth]{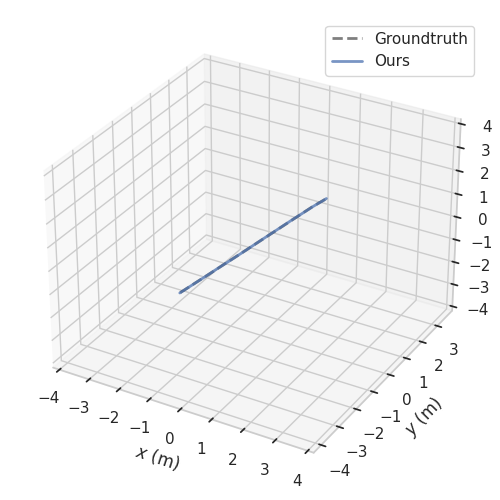}} \\

Sample frames & Tartan-VO~\cite{wang2020tartanvo} & DROID-SLAM~\cite{teed2021droid} & COLMAP~\cite{schonberger2016structure} & Ours \\

\end{tabular}
\caption{Qualitative results of moving camera localization on MPI Sintel dataset \cite{butler2012naturalistic}}
\label{fig::sintel}
\end{figure*}
\begin{table}[h]
\begin{center}
\setlength{\tabcolsep}{5.0pt}
\centering
\caption{Quantitative evaluation on MPI Sintel dataset \cite{butler2012naturalistic}. COLMAP \cite{schonberger2016structure} is additionally compared on its successful subset. Metrics are averaged across sequences}
\begin{tabular}{clccc}
\toprule
& Methods & ATE (m) & RPE trans (m) & RPE rot (deg)\\

\midrule
& COLMAP~\cite{schonberger2016structure} & 0.145 & 0.035 & 0.550\\
COLMAP & MAT~\cite{zhou2020motion} +~\cite{schonberger2016structure} & 0.069 & 0.024 & 0.726\\
subset & Mask-RCNN~\cite{he2017mask} +~\cite{schonberger2016structure} & 0.109 & 0.039 & 0.605\\
& Ours & \textbf{0.019} & \textbf{0.005} & \textbf{0.124} \\
\midrule
& COLMAP~\cite{schonberger2016structure} & X & X & X\\
& R-CVD~\cite{kopf2021robust} & 0.360 & 0.154 & 3.443 \\
Full set & Tartan-VO~\cite{wang2020tartanvo} & 0.290 & 0.092 & 1.303\\
& DROID-SLAM~\cite{teed2021droid} & 0.175 & 0.084 & 1.912\\
& Ours & \textbf{0.129} & \textbf{0.031} & \textbf{0.535}\\
\bottomrule
\end{tabular}
\label{tab::sintel}
\end{center}
\vspace{-10pt}
\end{table}

\subsection{Evaluation on MPI Sintel dataset}
MPI Sintel dataset~\cite{butler2012naturalistic} contains 23 synthetic sequences of highly dynamic scenes. We remove sequences that are not valid to evaluate monocular camera pose (e.g. static cameras, perfectly straight line), resulting in a total of 14 sequences for comparison. We compare our system with both feature based indirect SfM method COLMAP~\cite{schonberger2016structure} and state-of-the-art deep learning methods~\cite{kopf2021robust, wang2020tartanvo, teed2021droid}. 
The results are summarized in Table~\ref{tab::sintel}. We also provide comparisons with representative SLAM methods ORB-SLAM~\cite{mur2015orb} and DynaSLAM~\cite{bescos2018dynaslam} in the supplementary materials. Since COLMAP fails in 5 of 14 sequences, we perform comparison on the subset of other 9 sequences. Furthermore, we set up baselines by extracting motion masks from state-of-the-art methods MAT~\cite{zhou2020motion} and Mask-RCNN~\cite{he2017mask} to augment COLMAP \cite{schonberger2016structure}, where no feature points are extracted in the dynamic region. For Mask-RCNN \cite{he2017mask}, all the pixels that belong to potentially dynamic objects (person, vehicle, animals) are considered dynamic. As shown in Table \ref{tab::sintel}, while explicit motion removal improves the perforamance, our method outperforms compared baselines by a largin margin thanks to the advantages of long-range point trajectories. For learning-based methods, Tartan-VO~\cite{wang2020tartanvo} and DROID-SLAM~\cite{teed2021droid} are both trained on large-scale dataset TartanAir~\cite{wang2020tartanair} and have demonstrated strong generalization ability across different datasets. However, they struggle on dynamic scenes and fail to predict reliable camera trajectories. Some qualitative examples are shown in Figure \ref{fig::sintel}, where our system produces accurate camera poses over highly dynamic sequences.

\subsection{Evaluation on ScanNet}
\begin{figure}[tb]
\begin{center}
\scriptsize
\begin{tabular}{ccccc}
\centering
{\includegraphics[width=0.17\linewidth]{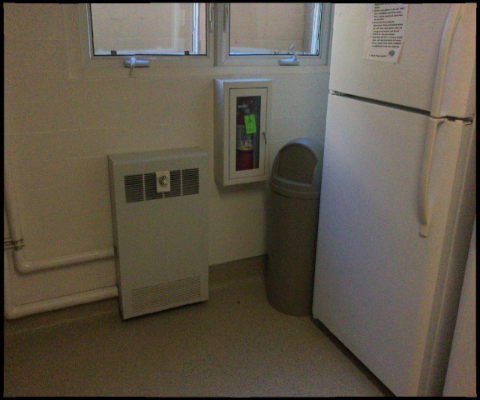}} & 
{\includegraphics[width=0.17\linewidth]{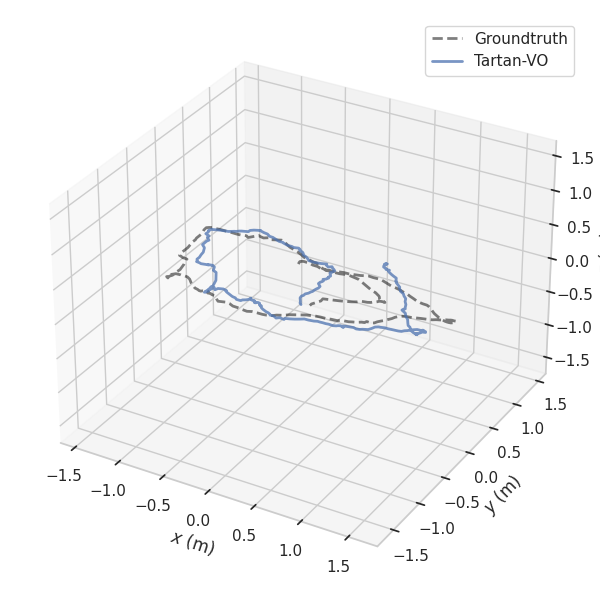}} & 
{\includegraphics[width=0.17\linewidth]{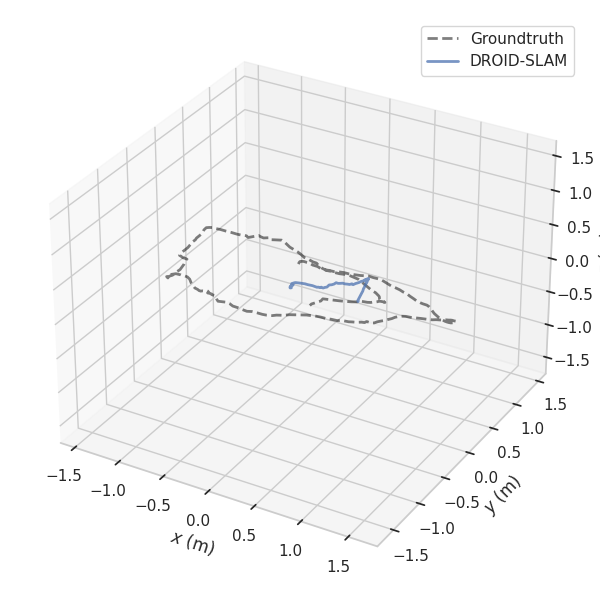}} &
{\includegraphics[width=0.17\linewidth]{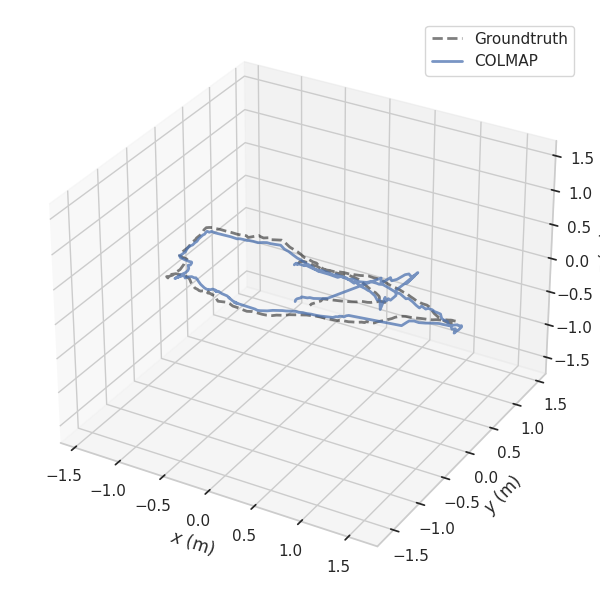}} & 
{\includegraphics[width=0.17\linewidth]{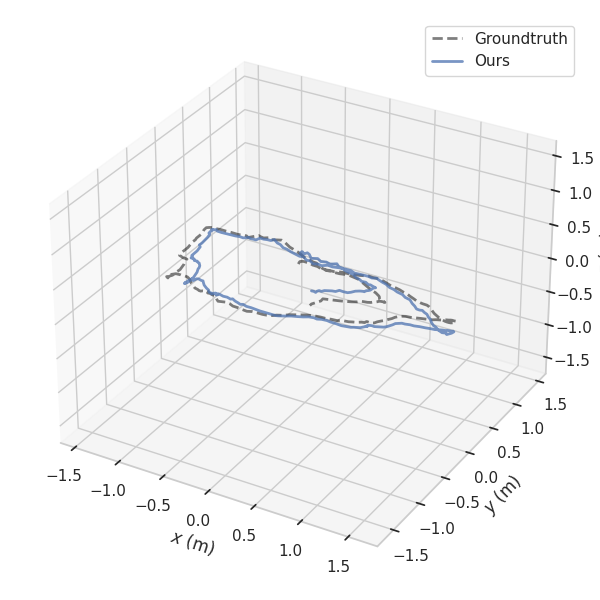}} \\

{\includegraphics[width=0.17\linewidth]{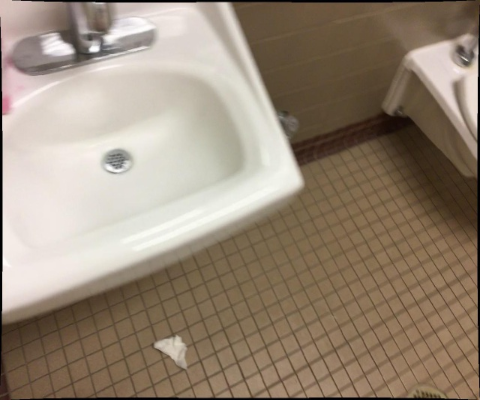}} & 
{\includegraphics[width=0.17\linewidth]{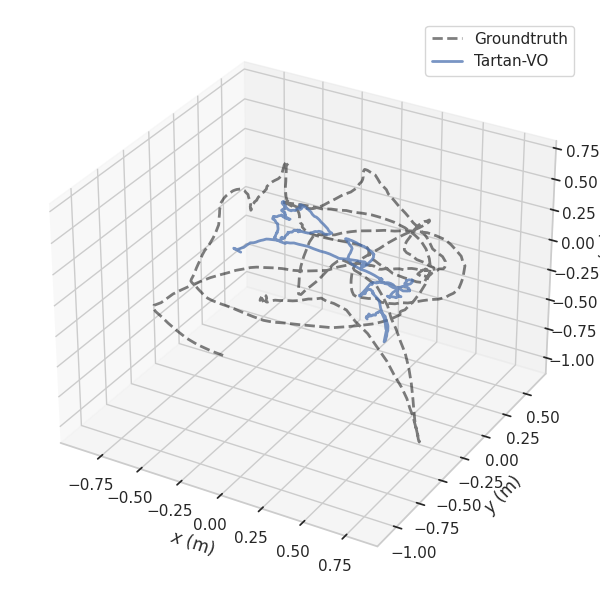}} & 
{\includegraphics[width=0.17\linewidth]{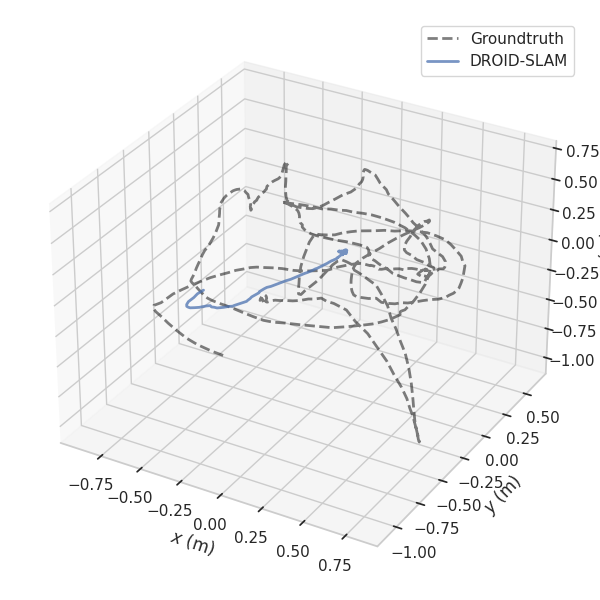}} & 
{\includegraphics[width=0.17\linewidth]{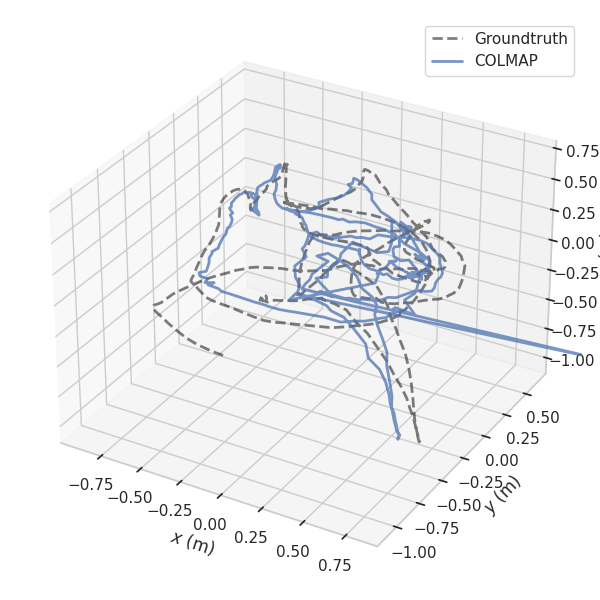}} &
{\includegraphics[width=0.17\linewidth]{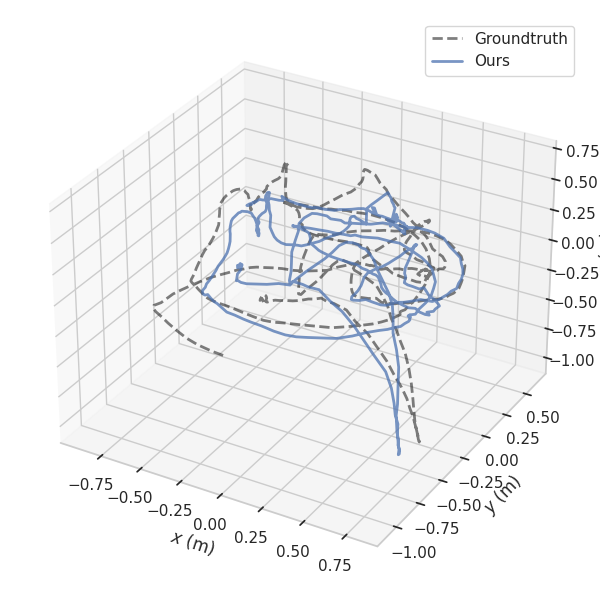}} \\

Sample frame & Tartan-VO~\cite{wang2020tartanvo} & DROID-SLAM~\cite{teed2021droid} & COLMAP~\cite{schonberger2016structure} & Ours \\

\end{tabular}
\caption{Qualitative results of moving camera localization on ScanNet dataset \cite{dai2017scannet}}
\label{fig::scannet}
\vspace{-5pt}
\end{center}
\end{figure}
\begin{table}[h]
\setlength{\tabcolsep}{5.0pt}
    \centering
    \caption{Quantitative evaluation on ScanNet \cite{dai2017scannet}. COLMAP \cite{schonberger2016structure} is additionally compared on its successful subset. Metrics are averaged across sequences}
    \begin{tabular}{clccc}
    \toprule
        & Methods & ATE (m) & RPE trans (m) & RPE rot (deg)\\
    
    \midrule
        COLMAP & COLMAP~\cite{schonberger2016structure} & \textbf{0.171} & 0.064 & 2.900\\
        subset & Ours & 0.319 & \textbf{0.017} & \textbf{0.632} \\
        \midrule
        & COLMAP~\cite{schonberger2016structure} & X & X & X\\
        & R-CVD~\cite{kopf2021robust} & 0.468 & 0.065 & 7.626 \\
        Full set & Tartan-VO~\cite{wang2020tartanvo} & 0.353 & 0.045 & 2.620\\
        & DROID-SLAM~\cite{teed2021droid} & 0.687 & 0.038 & 3.117\\
        & Ours & \textbf{0.349} & \textbf{0.024} & \textbf{0.924}\\
        \toprule
        
    \end{tabular}
    \label{tab::scannet}
\vspace{-10pt}
\end{table}
To study the generalization of the proposed dense indirect SfM system, we further test our method on fully static indoor dataset ScanNet~\cite{dai2017scannet}. Some sequences in ScanNet contain inf values in groundtruth poses and we exclude them from evaluation, resulting in out of 20 test sequences. The results are shown in Table~\ref{tab::scannet}. Our method consistently improve dense baselines and provide reasonable camera trajectories with complex camera motions, as shown in Figure~\ref{fig::scannet}. As for feature based method, COLMAP fails in 3 sequences out of 17, while our dense indirect system succeeds on all 17 sequences. We further present comparisons on the successful subset of COLMAP. Our method is slightly behind on ATE, but better on RPEs. This is possibly due to the degraded quality of indoor optical flow and the missing loop closure in our global bundle adjustment.

\subsection{Qualitative evaluation on in-the-wild videos}
\begin{figure*}[tb]
\begin{tabular}{cccccccc}
\centering
{\includegraphics[width=0.12\linewidth]{figures/imgs/in-the-wild/drift_straight_10.jpg}} & 
{\includegraphics[width=0.12\linewidth]{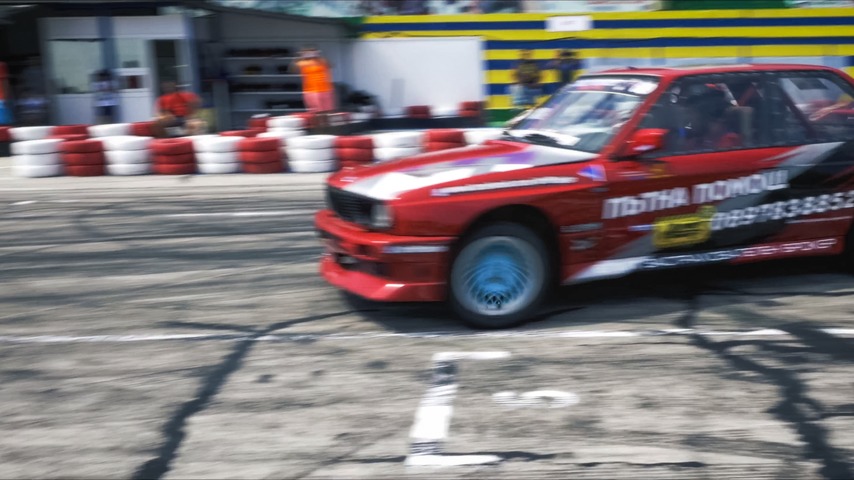}} &
{\includegraphics[width=0.12\linewidth]{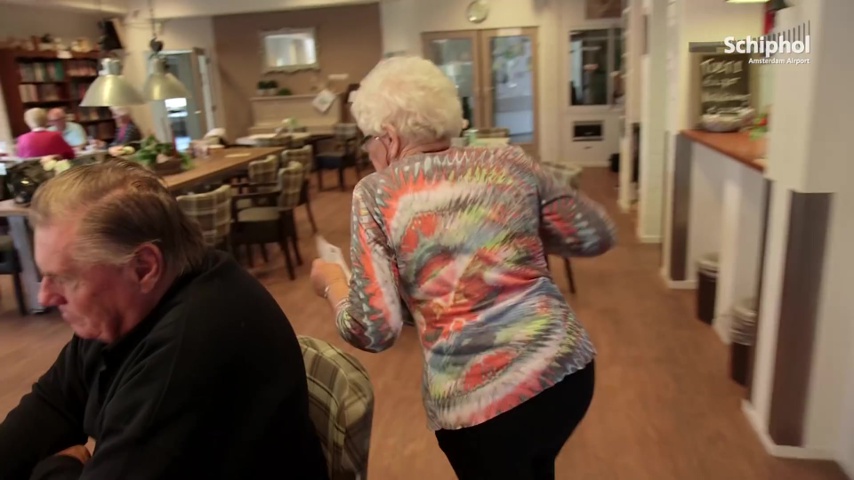}} & 
{\includegraphics[width=0.12\linewidth]{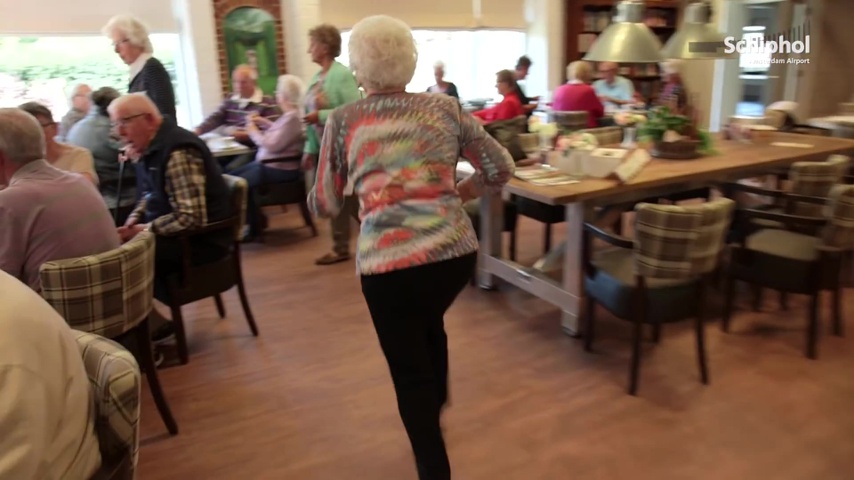}} &
{\includegraphics[width=0.12\linewidth]{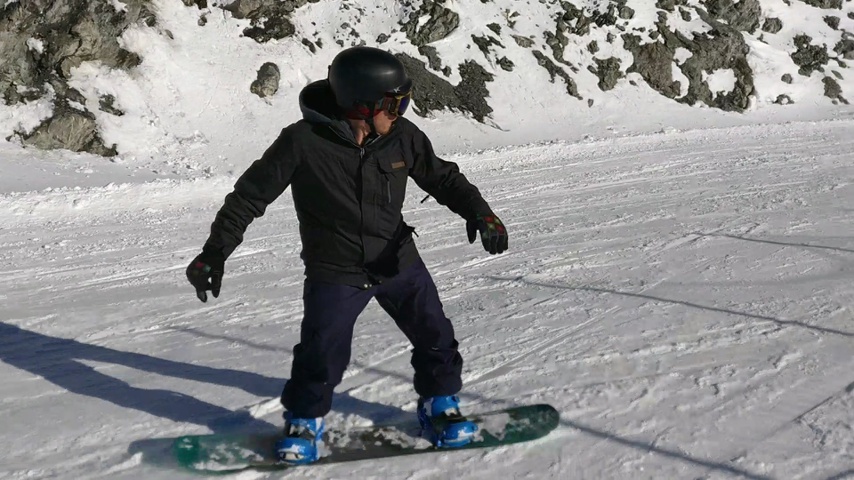}} &
{\includegraphics[width=0.12\linewidth]{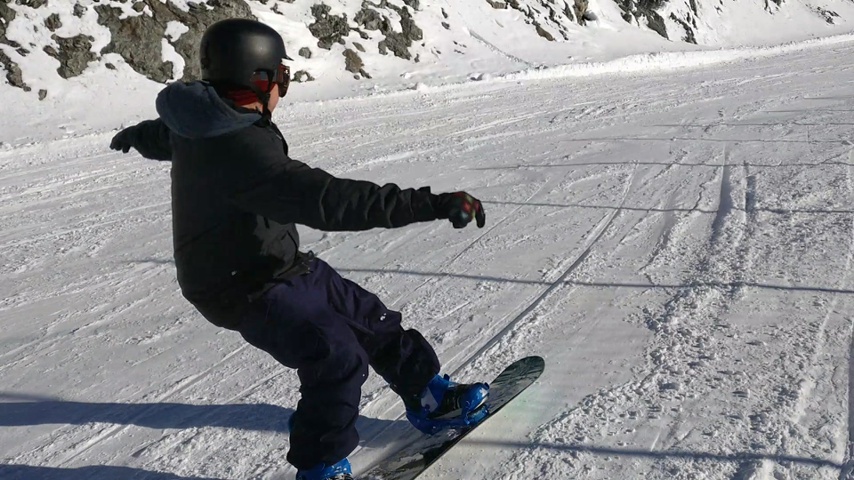}} &
{\includegraphics[width=0.12\linewidth]{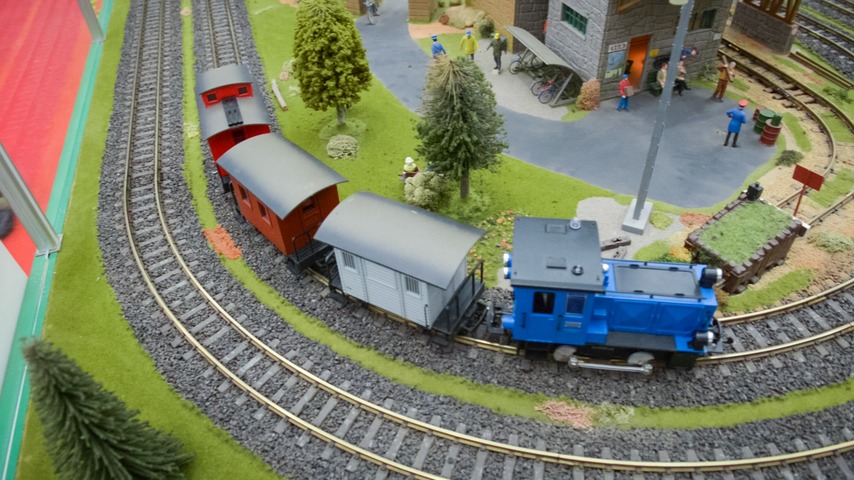}} &
{\includegraphics[width=0.12\linewidth]{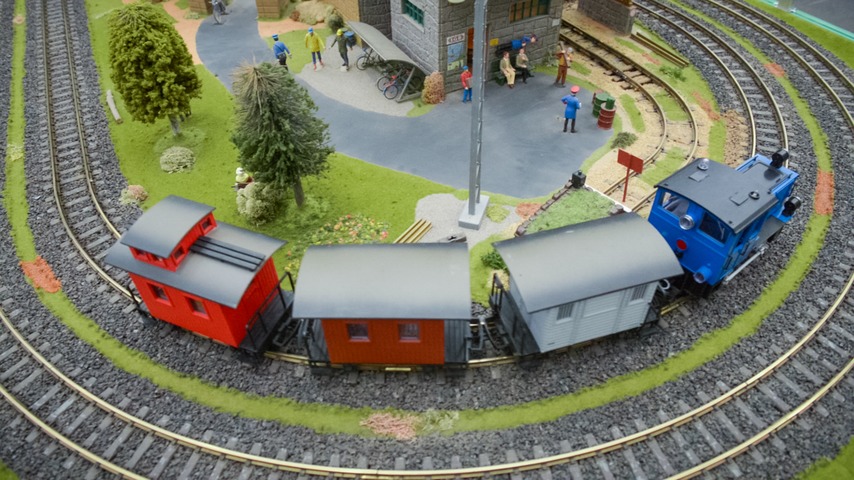}} \\


\multicolumn{2}{c}{\includegraphics[width=0.24\linewidth]{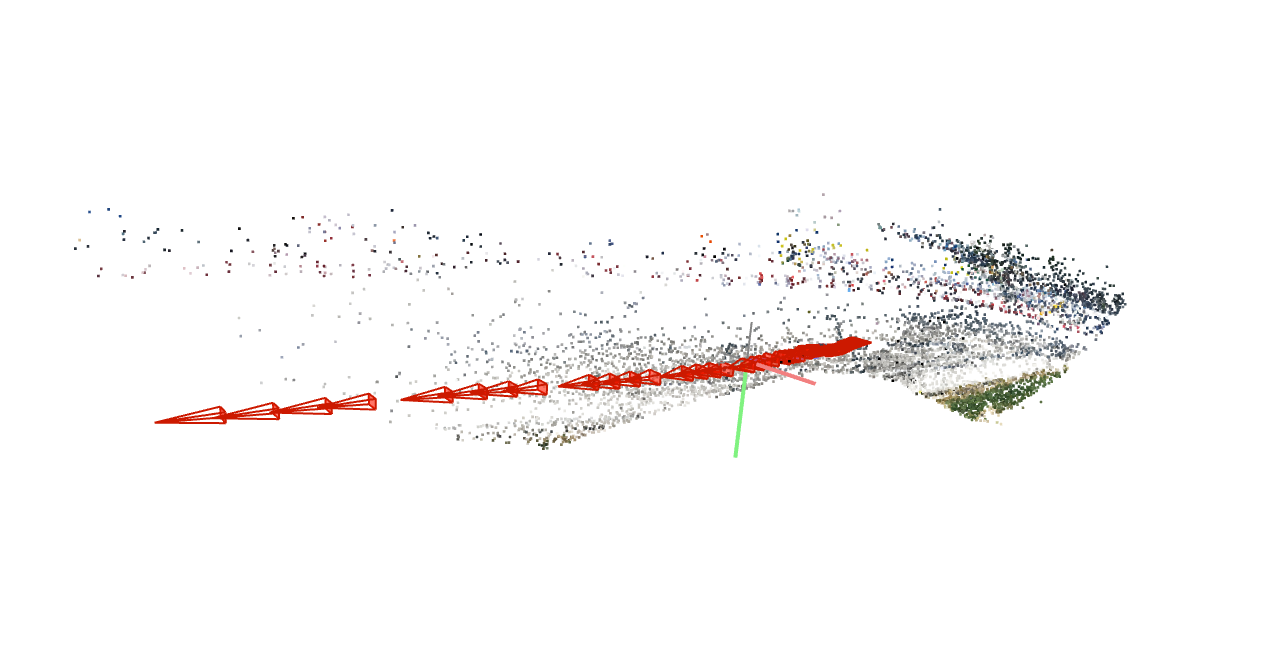}} & 
\multicolumn{2}{c}{{\includegraphics[width=0.24\linewidth]{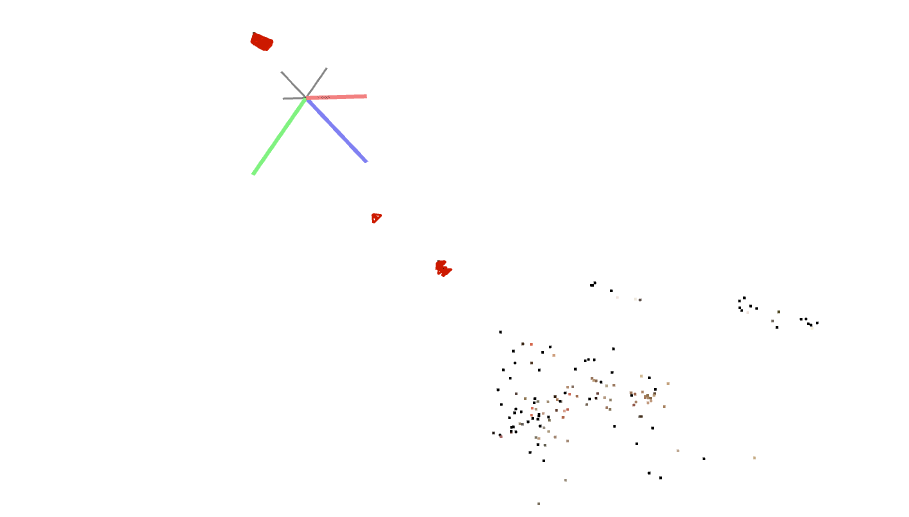}}} & 
\multicolumn{2}{c}{{\includegraphics[width=0.24\linewidth]{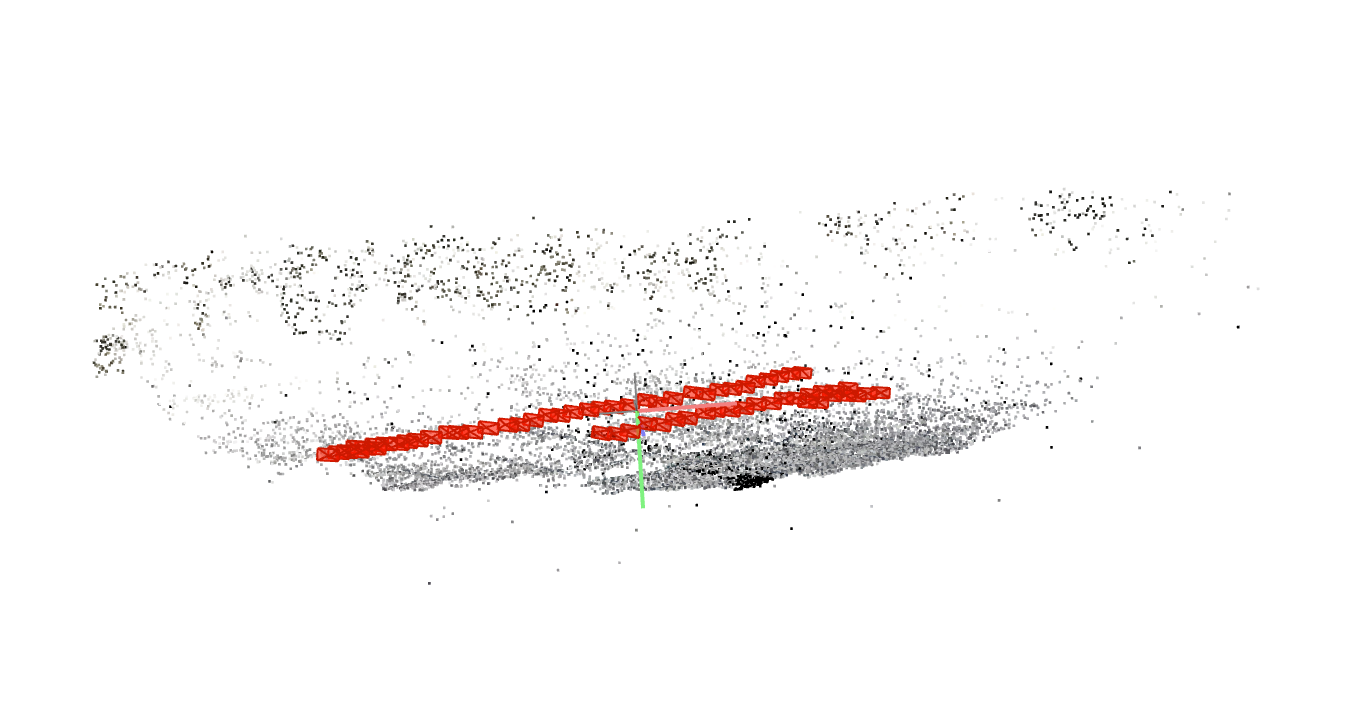}}} &
\multicolumn{2}{c}{{\includegraphics[width=0.24\linewidth]{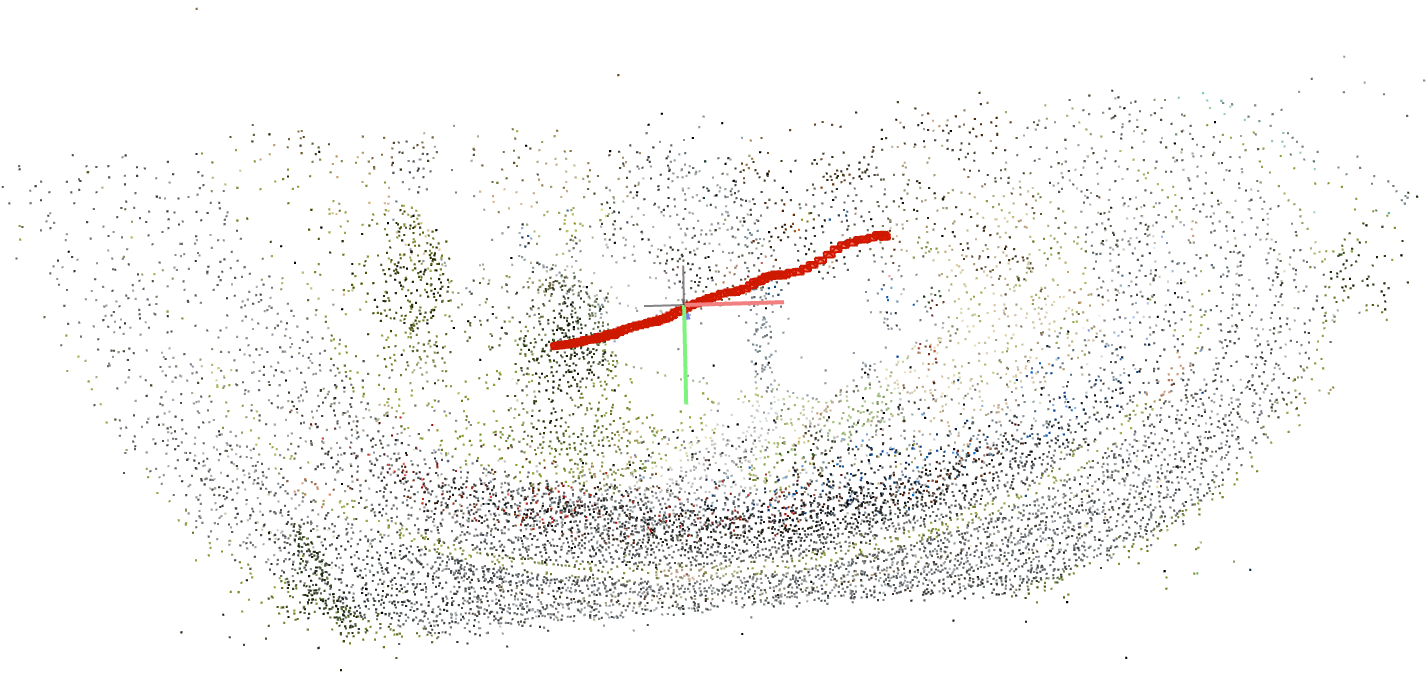}}} \\

\multicolumn{2}{c}{\includegraphics[width=0.24\linewidth]{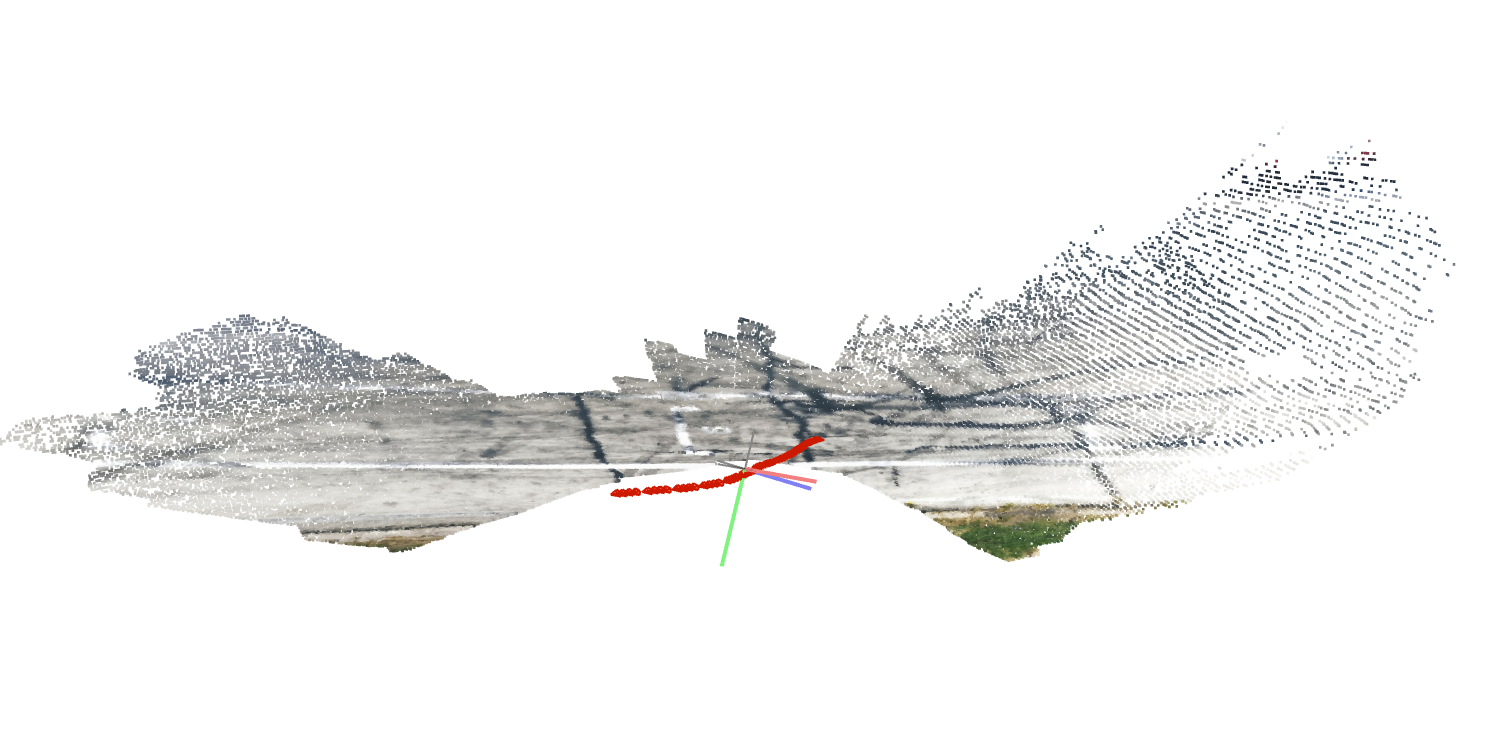}} & 
\multicolumn{2}{c}{{\includegraphics[width=0.24\linewidth]{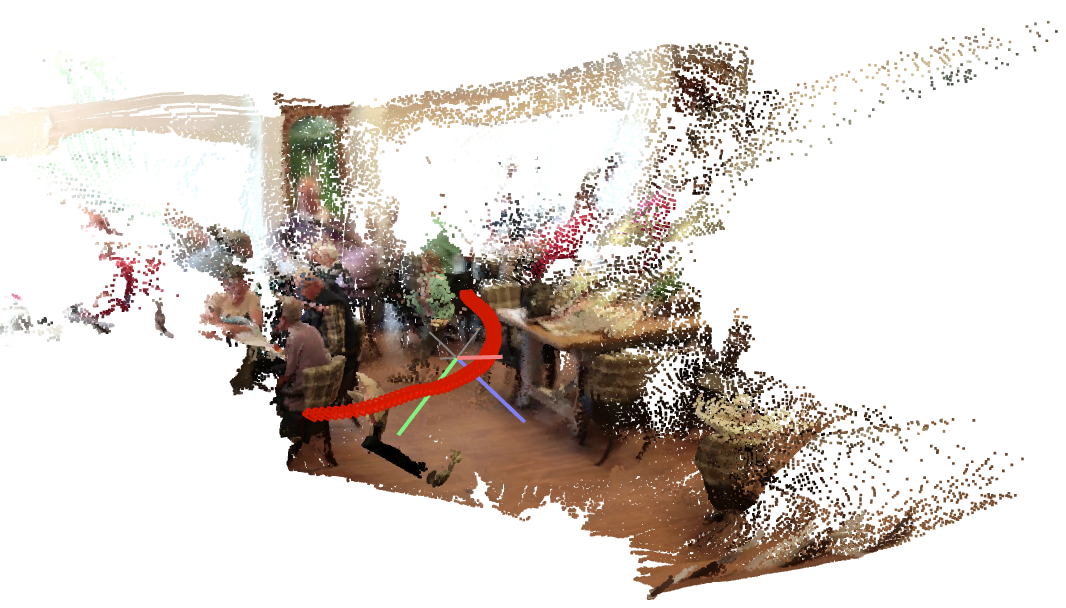}}} & 
\multicolumn{2}{c}{{\includegraphics[width=0.24\linewidth]{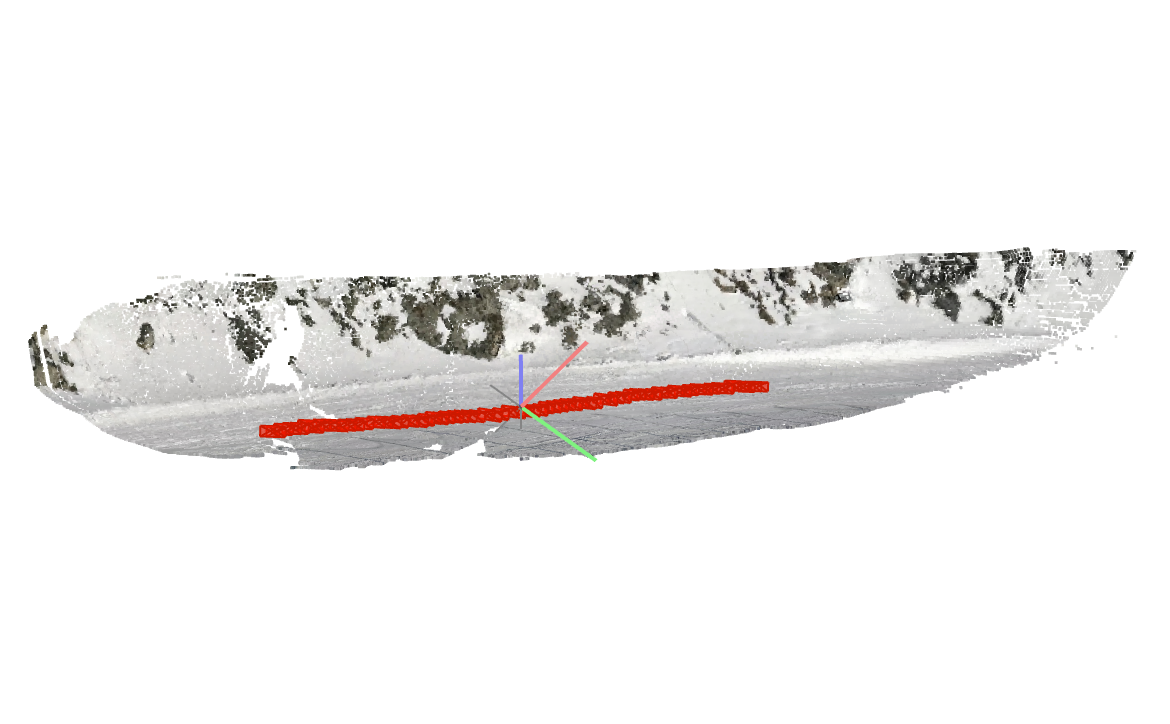}}} &
\multicolumn{2}{c}{{\includegraphics[width=0.24\linewidth]{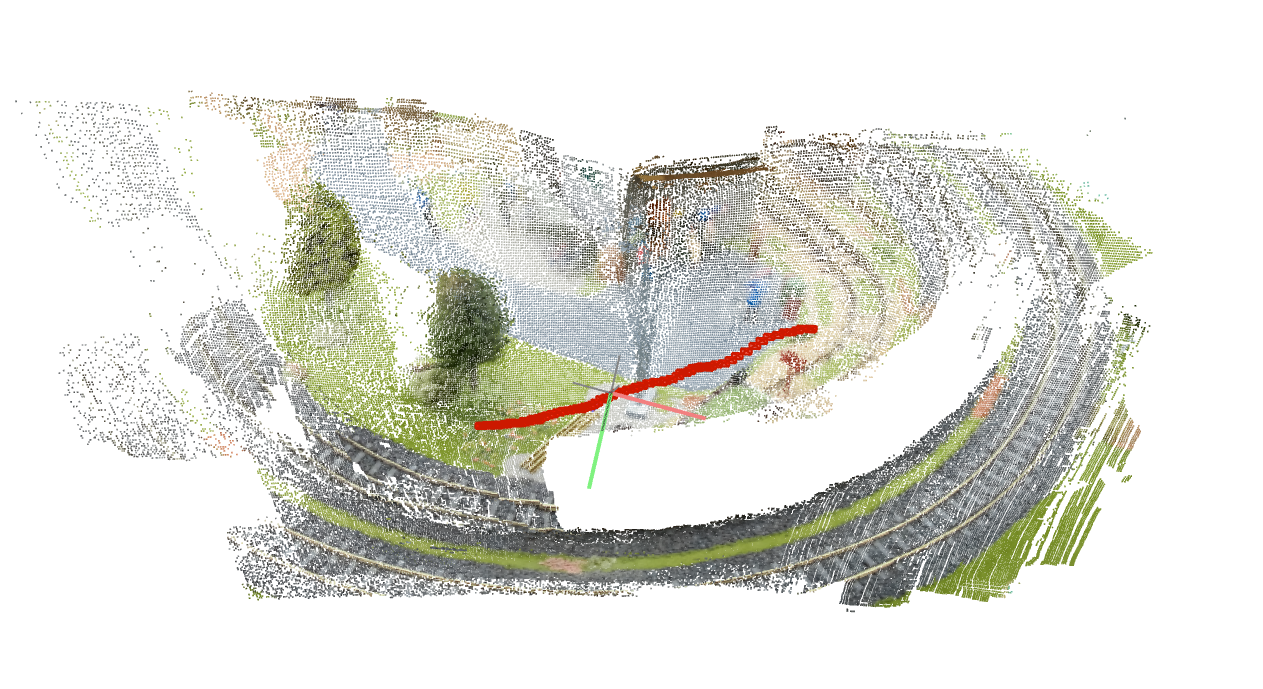}}} \\

\end{tabular}

\caption{\textbf{Top row:} COLMAP \cite{schonberger2016structure}. \textbf{Bottom row:} Ours. Qualitative comparison on in-the-wild videos from DAVIS \cite{perazzi2016benchmark}. Our dense indirect system produces robust camera trajectory and denser maps for videos with complex foreground motion}
\label{fig::wild}
\end{figure*}
As our system is general, we further present results and comparisons with COLMAP \cite{schonberger2016structure} on in-the-wild monocular videos with complex motion of dynamic objects in Figure~\ref{fig::wild}. Videos are taken from DAVIS~\cite{perazzi2016benchmark}. While COLMAP sometimes fails to predict reasonable camera poses (see the first and second sample of Figure~\ref{fig::wild}), our method generalizes well in the wild, and is more robust to dynamic objects. Furthermore, our dense indirect solution is able to build significantly denser 3d reconstructions, which demonstrates the potential of indirect methods for recovering fine-grained 3D geometry.

\subsection{Ablation Study}
\begin{table}[t]
\setlength{\tabcolsep}{2.0pt}
    \centering
    \caption{Results of ablation studies on MPI Sintel dataset \cite{butler2012naturalistic}. We report metrics for all methods on both the full set (left) and the successful subset for SIFT + Global BA (right). ``Optim" denotes trajectory optimization and ``Seg" denotes trajectory-based motion segmentation and its use for global bundle adjustment}
    \begin{tabular}{lcccccccccccc}
    \toprule
        Methods && \multicolumn{3}{c}{ATE (m)} & & \multicolumn{3}{c}{RPE trans (m)} && \multicolumn{3}{c}{RPE rot (deg)}\\

    \midrule
        SIFT + Global BA &&  X &/& 0.060   &&  X &/& 0.042    &&  X &/& 0.635 \\
        SIFT + MAT~\cite{zhou2020motion} + Global BA &&   X &/& 0.054   &&  X &/& 0.055   &&   X &/& 0.621 \\
        Traj + Global BA && X &/& 0.071 &&   X &/& 0.041   &&  X &/& 0.969 \\
        Traj + Optim + Global BA &&  X &/& 0.072    &&    X &/& 0.042     &&    X &/& 0.929 \\
        Traj + Seg + Global BA && 0.146 &/& 0.046    &&    0.039 &/& 0.015    &&    0.567 &/& 0.212 \\
        Traj + Optim + Seg + Global BA && \textbf{0.129} &/& \textbf{0.042}    &&     \textbf{0.031} &/& \textbf{0.013}    &&     \textbf{0.535} &/& \textbf{0.199} \\
        \toprule
        
    \end{tabular}
    \label{tab::ablation}
\end{table}
We ablate different components of our system on MPI Sintel dataset \cite{butler2012naturalistic}. Table \ref{tab::ablation} shows the results. Without trajectory motion segmentation, all candidate correspondences, including those on moving objects, are used for map construction and global bundle adjustment. Results show that this leads to significantly degraded accuracy of recovered poses, indicating the important of removing dynamic pixels. Moreover, by optimizing point trajectories with path consistency, the mean endpoint error of trajectories is reduced by around 10\%, which consequently improves camera localization (the last two rows). To further validate the necessity of dense point trajectories, we introduce two feature-based baselines that are built on SIFT \cite{lowe2004distinctive}, where one is also integrated with MAT \cite{zhou2020motion} for removing moving pixels. As shown in Table~\ref{tab::ablation},  these two methods sometimes fail on sequences, and are behind on accuracy when tested on their successful subset compared to our system, demonstrating the effectiveness of dense point trajectories.

\begin{figure*}[tb]
\centering
\begin{tabular}{cccc}

{\includegraphics[width=0.22\linewidth]{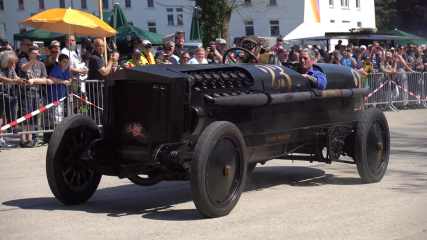}} & 
{\includegraphics[width=0.22\linewidth]{figures/imgs/motion_seg/m2_davis_dance73-img.png}} & 
{\includegraphics[width=0.22\linewidth]{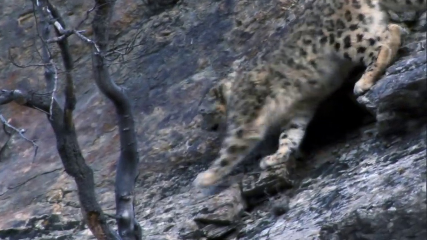}} &
{\includegraphics[width=0.22\linewidth]{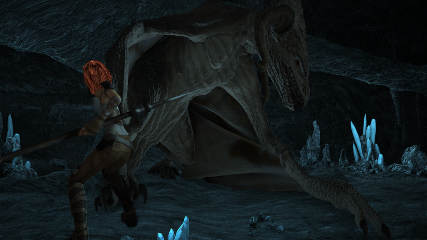}} \\

{\includegraphics[width=0.22\linewidth]{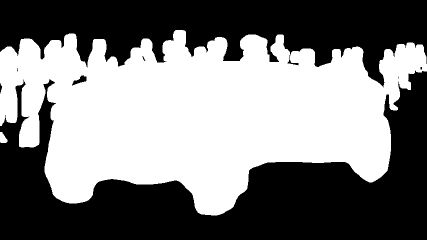}} & 
{\includegraphics[width=0.22\linewidth]{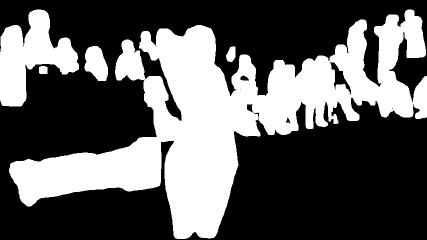}} & 
{\includegraphics[width=0.22\linewidth]{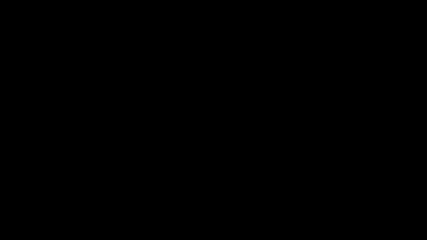}} &
{\includegraphics[width=0.22\linewidth]{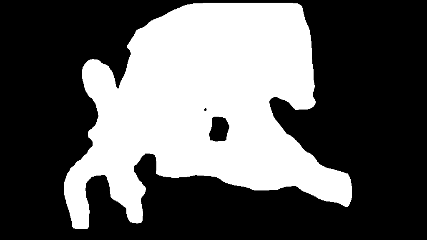}} \\

{\includegraphics[width=0.22\linewidth]{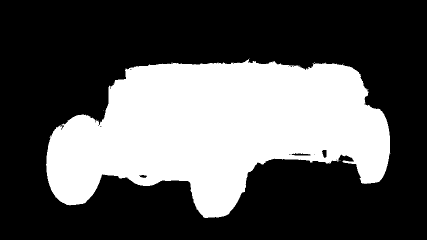}} & 
{\includegraphics[width=0.22\linewidth]{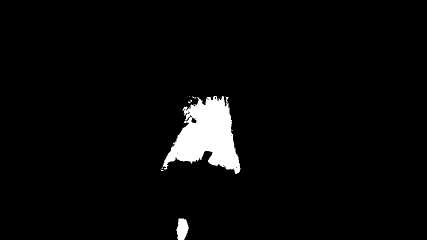}} & 
{\includegraphics[width=0.22\linewidth]{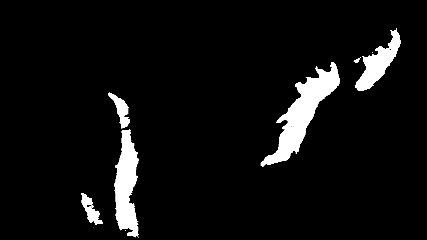}} &
{\includegraphics[width=0.22\linewidth]{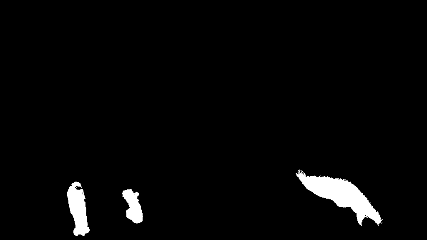}} \\

{\includegraphics[width=0.22\linewidth]{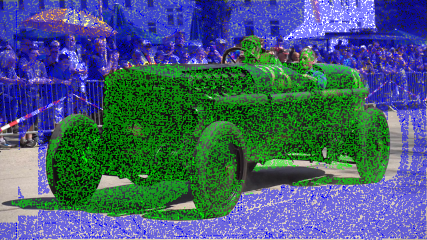}} & 
{\includegraphics[width=0.22\linewidth]{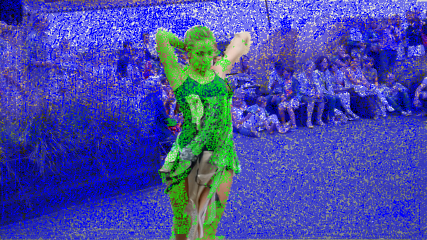}} & 
{\includegraphics[width=0.22\linewidth]{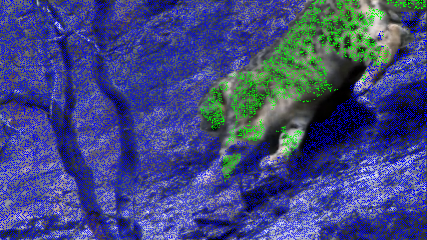}} &
{\includegraphics[width=0.22\linewidth]{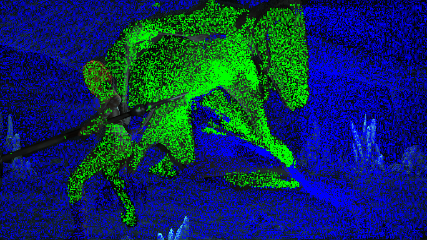}} \\

\end{tabular}
\caption{Qualitative results on motion segmentation. \textbf{From top to bottom:} sample image, Mask-RCNN \cite{he2017mask}, MAT \cite{zhou2020motion} and our trajectory-based motion segmentation}
\label{fig::motion_seg}
\end{figure*}
\begin{table}[t]
\setlength{\tabcolsep}{5.0pt}
    \centering
    \caption{Evaluation and ablation studies for motion segmentation on MPI Sintel dataset \cite{butler2012naturalistic}. We compare our method with several state-of-the-arts and ablate different components in our design on trajectory-based motion segmentation}
    \begin{tabular}{lcccc}
    \toprule
        Methods & mIoU (\%) & Precision & Recall & F1-score \\
    \midrule
       MAT~\cite{zhou2020motion} & 47.5 & \textbf{0.82} & 0.54 & 0.56 \\
       COS~\cite{lu2020_pami} & 55.0 & 0.67 & \textbf{0.77} & 0.65 \\
       MotionGrouping~\cite{yang2021self} & 16.2 & 0.64 & 0.19 & 0.25 \\
       AMD~\cite{liu2021emergence} & 31.5 & 0.42 & 0.62 & 0.45 \\
       \midrule
       Two-branch CNN w/o depth & 29.2 & 0.54 & 0.49 & 0.39\\
       Two-branch CNN & 33.7 & 0.62 & 0.50 & 0.44\\
       Ours with MLP encoder & 54.8 & 0.67 & 0.73 & 0.66\\
       Ours with PointNet decoder & 46.3 & 0.65 & 0.67 & 0.58\\
       Ours w/o depth & 54.6 & 0.72 & 0.70 & 0.65\\
       \midrule
       Ours & \textbf{60.6} & 0.79 & 0.74 & \textbf{0.72}\\
        \toprule
    
    \end{tabular}
    \label{tab::motion_ablation}
\end{table}
To study the effects of different components in the trajectory motion segmentation network, we perform ablation studies on MPI Sintel dataset \cite{butler2012naturalistic} by evaluating segmentation w.r.t groundtruth motion masks. Since trajectory motion segmentation methods only offer predictions for trajectories, we map the motion labels into corresponding pixel locations and evaluate all methods only on these pixels. Table~\ref{tab::motion_ablation} shows the results. We compare our method with state-of-the-art supervised motion segmentation method \cite{zhou2020motion} that is trained on DAVIS~\cite{perazzi2016benchmark}
and YouTube-VOS~\cite{xu2018youtube}. Our method, while only trained on synthetic FlyingThings3D dataset~\cite{mayer2016large}, achieves better motion segmentation even without depth information. Figure \ref{fig::motion_seg} shows some qualitative results, where our method predicts reliable motion masks for moving pixels. To further support the novel design of the network, we introduce a baseline that utilize two-branch CNN networks with ResNet-50~\cite{he2016deep} to aggregate motion, appearance, and optionally also depth information. However, results indicate that this two-branch network exhibits low generalization ability when trained on FlyingThings3D dataset. Finally, we also ablate the architecture design by substituting the encoder with vanilla MLPs and the decoder with PointNet \cite{qi2017pointnet}. Results clearly show the advantages of using attention-based encoder and local-global context-aware decoder in our system.

\paragraph{Runtime analysis.} For a 50-frame video, our system runs within 690s on average, including all the I/O time of each module. Specifically, optical flow extraction takes around 57s, the acquisition of dense point trajectories takes 200s, the trajectory motion segmentation takes around 65s, and the final global BA takes 368s. For comparison, COLMAP \cite{schonberger2016structure} takes 410s on average. Note that it is computationally unaffordable to run COLMAP with dense correspondences from the acquired dense point trajectories, while our method achieves dense correspondence with reasonable overhead.

\section{Conclusions}
In this work, we present a general \textit{dense indirect} system for localizing moving cameras from in-the-wild videos. The key to the success of our method is to optimize and exploit long-range dense video correspondences as point trajectories, which are used for motion analysis and global bundle adjustment. A specially designed trajectory-based motion segmentation network is proposed to process the irregular point trajectory data. Experimental results show that our method can recover reliable camera trajectories from in-the-wild videos with complex motion patterns. Possible future directions include advanced optimization for long-range point trajectories and integration of loop closure for pose optimization.
\\

\noindent
\textbf{Acknowledgements} We thank anonymous reviewers for their valuable feedback. This work was supported by the Natural Science Foundation of China (61725204) and Tsinghua University Initiative Scientific Research Program.  

\appendix
\section{More Implementation Details}
\paragraph{Point Trajectory.} We use the pretrained \textit{raft-things} model for RAFT~\cite{teed2020raft} in all our experiments. It is trained on FlyingChairs and FlyingThings3D~\cite{mayer2016large}. For optical flow forward-backward consistency check, we use the threshold of 1px for MPI Sintel dataset~\cite{butler2012naturalistic}, and 3px for ScanNet dataset~\cite{dai2017scannet}.

\paragraph{Motion Segmentation.} For training trajectory motion segmentation network,  we first prepare the trajectory data of FlyingThings3D dataset~\cite{mayer2016large}. The dataset consists of over 2000 training scenes, and each scene contains 10 video frames, together with groundtruth optical flow, camera parameters and depth maps. We run our dense point trajectory generation algorithm to track trajectories from groundtruth optical flows, and then calculate trajectory groundtruth motion labels by comparing optical flow with rigid flow from depths and camera poses. We infer the relative depth information by pretrained MiDaS~\cite{ranftl2019towards} model. For all experiments, we use the pretrained \textit{midas-v21} model. During training, we directly take all the point trajectories from 10 video frames and output the per-trajectory motion label. Weighted binary cross-entropy loss is then applied.  

\paragraph{Global Bundle Adjustment (BA). } 
The implementation of our pipeline is mainly based on the Theia SfM system~\cite{theia-manual}. With the dense correspondences sampled from the point trajectories, we first compute two-view geometry \cite{hartley1997defense} between valid image pairs and decompose the relative poses. In particular, the view pairs with very few or extremely noisy correspondences are detected with geometric verification \cite{schonberger2016structure} and ignored in the subsequent stages. Then, L1-IRLS rotation averaging \cite{chatterjee2013efficient} is applied to estimate the global orientations from the relative rotations among those valid pairs. After filtering outlier pairs with large errors on the relative rotations, we solve for the relative translations with the global rotations and apply LUD translation averaging \cite{ozyesil2015robust} to get global translations. With these initial global poses, we incrementally triangulate 2D observations as in \cite{schonberger2016structure} and perform bundle adjustment on all the poses and 3D points to get the final output camera poses. Note that since we triangulate over the correspondences directly sampled from the dense point trajectories, each constraint in the bundle adjustment exactly corresponds to a part of the original point trajectory with geometric filtering, enabling effective global bundle adjustment over the input trajectory observations.

\section{Comparisons with SLAM methods}
We provide the quantitative comparison results on Sintel dataset with representative monocular SLAM methods ORB-SLAM~\cite{mur2015orb} and DynaSLAM~\cite{bescos2018dynaslam}. ORB-SLAM~\cite{mur2015orb} is the most popular feature-based monocular SLAM method which utilizes robust front-end tracking and local bundle-adjustment to achieve accurate camera localization. Built on top of ORB-SLAM, DynaSLAM~\cite{bescos2018dynaslam} further introduces semantics to remove potentially dynamic objects to improve the robustness. Since ORB-SLAM and DynaSLAM only provide key-frame poses, we compared the localization accuracy solely on their key-frames. For Sintel dataset, both ORB-SLAM and DynaSLAM \textbf{consistently fail in 8 of 14 sequences}, and we summarize the results from other 6 successful sequences in Table~\ref{tab::supp_slam}. Our method surpasses both of them by a large margin even on their successful subset. Furthermore, ORB-SLAM and DynaSLAM \textbf{fail on all 17 ScanNet sequences}, probably due to large motion blur and poorly textured regions, while our method consistently provides reasonable camera poses.
\begin{table}[h]
\vspace{-5pt}
\begin{center}
\centering
\caption{Evaluation on the successful subset (6 out of 14 sequences) of ORB-SLAM / DynaSLAM on MPI Sintel dataset.}

\begin{tabular}{lccc}
\toprule
Methods & ATE (m) & RPE trans (m) & RPE rot (deg)\\

\midrule
ORB-SLAM & 0.042 & 0.022 & 0.402\\
Ours & \textbf{0.009} & \textbf{0.006} & \textbf{0.101} \\
\midrule
DynaSLAM & 0.020 & 0.019 & 0.359\\
Ours & \textbf{0.007} & \textbf{0.005} & \textbf{0.090}\\
\bottomrule
\end{tabular}
\label{tab::supp_slam}
\end{center}
\vspace{-5pt}
\end{table}

\section{Per-scene Results on MPI Sintel and ScanNet}
We show the per-scene comparison results of MPI Sintel~\cite{butler2012naturalistic} and ScanNet~\cite{dai2017scannet} dataset in Table~\ref{tab::sintel_perscene} and Table~\ref{tab::scannet_perscene}. For MPI Sintel, our method achieves the best performance on most sequences, demonstrating the advantage of the proposed system in dynamic scenarios. For fully static indoor dataset ScanNet, our method retain comparable performance with COLMAP~\cite{schonberger2016structure}, slightly behind on ATE and better on RPEs. COLMAP globally matches the feature points between image pairs, thus naturally has the ability of loop closure, while our method lacks as point trajectories are accumulated sequentially. Although our method could achieve good relative pose estimations, the trajectory error will be accumulated without loop closure. This is possibly the main reason why our method is worse than COLMAP in ATE but better in RPEs. In the future, we aim to implement the loop closure inside our system by matching trajectories across frames.
\begin{table}[h]
\footnotesize
\begin{center}
\setlength{\tabcolsep}{3.0pt}
\centering
\caption{Per-scene results on MPI Sintel dataset \cite{butler2012naturalistic}.}
\scalebox{0.6}{
\begin{tabular}{cl|c|c|c|c|c|c|c}
\toprule
& Metrics & COLMAP~\cite{schonberger2016structure} & MAT~\cite{zhou2020motion} +~\cite{schonberger2016structure} & Mask-RCNN~\cite{he2017mask} +~\cite{schonberger2016structure} & R-CVD~\cite{kopf2021robust} & Tartan-VO~\cite{wang2020tartanvo} & DROID-SLAM~\cite{teed2021droid} & Ours\\
\midrule
\multirow{3}{*}{\textbf{alley\_2}} & ATE (m) & 0.072 & \textbf{0.002} & 0.072 & 0.026 & 0.062 & 0.057 & \textbf{0.002} \\
 & RPE trans (m) & 0.039 & 0.0009 & 0.038 & 0.056 & 0.049 & 0.035 & \textbf{0.0007} \\
 & RPE rot (deg) & 0.678 & 0.014 & 0.679 & 0.821 & 0.856 & 1.047 & \textbf{0.009} \\
\midrule

\multirow{3}{*}{\textbf{ambush\_4}} & ATE (m) & 0.030 & 0.174 & \textbf{0.029} & 0.171 & 0.100 & 0.104 & 0.068 \\
 & RPE trans (m) & 0.032 & 0.046 & 0.045 & 0.048 & 0.038 & 0.035 & \textbf{0.027} \\
 & RPE rot (deg) & 0.377 & 3.425 & 0.541 & 3.025 & 1.320 & 1.385 & \textbf{0.473} \\
\midrule

\multirow{3}{*}{\textbf{ambush\_5}} & ATE (m) & 0.028 & 0.090 & 0.004 & 0.230 & 0.098 & 0.112 & \textbf{0.002} \\
 & RPE trans (m) & 0.015 & 0.036 & 0.005 & 0.046 & 0.037 & 0.029 & \textbf{0.001} \\
 & RPE rot (deg) & 0.607 & 0.817 & 0.204 & 4.105 & 1.107 & 1.580 & \textbf{0.055} \\
\midrule

\multirow{3}{*}{\textbf{ambush\_6}} & ATE (m) & X & X & X & \textbf{0.199} & 0.205 & 0.289 & 0.269 \\
 & RPE trans (m) & X & X & X & 0.112 & 0.107 & \textbf{0.078} & 0.081 \\
 & RPE rot (deg) & X & X & X & 4.147 & 4.293 & 4.596 & \textbf{0.951} \\
\midrule

\multirow{3}{*}{\textbf{cave\_2}} & ATE (m) & X & X & X & \textbf{0.596} & 1.167 & 0.351 & 0.961 \\
 & RPE trans (m) & X & X & X & 0.171 & \textbf{0.131} & 0.172 & 0.142 \\
 & RPE rot (deg) & X & X & X & 7.508 & 4.112 & 5.489 & \textbf{3.678} \\
\midrule

\multirow{3}{*}{\textbf{cave\_4}} & ATE (m) & 0.051 & 0.049 & \textbf{0.044} & 0.179 & 0.120 & 0.155 & 0.068 \\
 & RPE trans (m) & 0.013 & 0.028 & 0.040 & 0.087 & 0.039 & 0.035 & \textbf{0.012} \\
 & RPE rot (deg) & 0.451 & 0.700 & 0.600 & 2.040 & 1.327 & 2.710 & \textbf{0.409} \\
\midrule

\multirow{3}{*}{\textbf{market\_2}} & ATE (m) & X & X & X & 0.032 & 0.068 & 0.011 & \textbf{0.003} \\
 & RPE trans (m) & X & X & X & 0.018 & 0.007 & \textbf{0.006} & 0.010 \\
 & RPE rot (deg) & X & X & X & 0.141 & 0.090 & \textbf{0.036} & 0.041 \\
\midrule

\multirow{3}{*}{\textbf{market\_5}} & ATE (m) & 1.105 & 0.284 & 0.816 & 1.213 & 1.158 & 0.912 & \textbf{0.012} \\
 & RPE trans (m) & 0.210 & 0.095 & 0.212 & 0.762 & 0.294 & 0.293 & \textbf{0.006} \\
 & RPE rot (deg) & 2.232 & 0.055 & 2.380 & 1.863 & 1.100 & 3.334 & \textbf{0.034} \\
\midrule

\multirow{3}{*}{\textbf{market\_6}} & ATE (m) & X & X & X & 0.248 & 0.260 & 0.057 & \textbf{0.018} \\
 & RPE trans (m) & X & X & X & 0.214 & 0.110 & 0.037 & \textbf{0.007} \\
 & RPE rot (deg) & X & X & X & 0.817 & 1.287 & 1.296 & \textbf{0.120} \\
\midrule

\multirow{3}{*}{\textbf{shaman\_3}} & ATE (m) & 0.012 & 0.006 & 0.009 & 0.054 & 0.008 & 0.001 & \textbf{0.0005} \\
 & RPE trans (m) & 0.003 & 0.005 & 0.007 & 0.023 & 0.006 & 0.002 & \textbf{0.0004} \\
 & RPE rot (deg) & 0.537 & 0.978 & 0.977 & 0.718 & 0.185 & 0.199 & \textbf{0.072} \\
\midrule

\multirow{3}{*}{\textbf{sleeping\_1}} & ATE (m) & \textbf{0.008} & 0.013 & \textbf{0.008} & 0.029 & 0.017 & 0.011 & \textbf{0.008} \\
 & RPE trans (m) & \textbf{0.001} & 0.006 & \textbf{0.001} & 0.019 & 0.011 & 0.006 & \textbf{0.001} \\
 & RPE rot (deg) & 0.053 & 0.530 & 0.046 & 0.668 & 0.344 & 0.479 & \textbf{0.042} \\
\midrule

\multirow{3}{*}{\textbf{sleeping\_2}} & ATE (m) & \textbf{0.0002} & \textbf{0.0002} & \textbf{0.0002} & 0.043 & 0.013 & 0.005 & 0.0007 \\
 & RPE trans (m) & \textbf{0.0002} & \textbf{0.0002} & \textbf{0.0002} & 0.049 & 0.022 & 0.0177 & \textbf{0.0002} \\
 & RPE rot (deg) & 0.008 & \textbf{0.006} & 0.007 & 0.446 & 0.267 & 0.139 & \textbf{0.006} \\
\midrule

\multirow{3}{*}{\textbf{temple\_2}} & ATE (m) & \textbf{0.004} & 0.006 & 0.006 & 1.245 & 0.447 & 0.073 & 0.011 \\
 & RPE trans (m) & 0.003 & \textbf{0.002} & \textbf{0.002} & 0.394 & 0.324 & 0.348 & \textbf{0.002} \\
 & RPE rot (deg) & 0.012 & \textbf{0.007} & 0.013 & 1.318 & 0.789 & 1.298 & 0.019 \\
\midrule

\multirow{3}{*}{\textbf{temple\_3}} & ATE (m) & X & X & X & 0.769 & 0.331 & \textbf{0.310} & 0.381 \\
 & RPE trans (m) & X & X & X & 0.161 & 0.105 & \textbf{0.093} & 0.149 \\
 & RPE rot (deg) & X & X & X & 20.592 & \textbf{1.166} & 3.230 & 1.583 \\
\bottomrule

\end{tabular}}
\label{tab::sintel_perscene}
\end{center}
\end{table}
\begin{table}[h]
\footnotesize
\begin{center}
\setlength{\tabcolsep}{3.0pt}
\centering
\caption{Per-scene results on ScanNet dataset \cite{dai2017scannet}.}
\scalebox{0.7}{
\begin{tabular}{cl|c|c|c|c|c}
\toprule
& Metrics & COLMAP~\cite{schonberger2016structure} & R-CVD~\cite{kopf2021robust} & Tartan-VO~\cite{wang2020tartanvo} & DROID-SLAM~\cite{teed2021droid} & Ours\\
\midrule
\multirow{3}{*}{\textbf{scene0707\_00}} & ATE (m) & \textbf{0.147} & 0.442 & 0.418 & 0.978 & 0.199 \\
 & RPE trans (m) & 0.059  & 0.092 & 0.065 & 0.043 & \textbf{0.020} \\
 & RPE rot (deg) & 0.803  & 7.301 & 2.914 & 3.530 & \textbf{0.574} \\
\midrule

\multirow{3}{*}{\textbf{scene0709\_00}} & ATE (m) & \textbf{0.143}  & 0.437 & 0.202 & 0.872 & 0.220 \\
 & RPE trans (m) & 0.067  & 0.088 & 0.063 & 0.052 & \textbf{0.015} \\
 & RPE rot (deg) & 0.780  & 6.852 & 2.698 & 3.187 & \textbf{0.523} \\
\midrule

\multirow{3}{*}{\textbf{scene0710\_00}} & ATE (m) & \textbf{0.073}  & 0.429 & 0.306 & 0.631 & 0.247 \\
 & RPE trans (m) & 0.016  & 0.039 & 0.027 & 0.030 & \textbf{0.012} \\
 & RPE rot (deg) & \textbf{0.371}  & 4.412 & 1.961 & 2.153 & 0.424 \\
\midrule

\multirow{3}{*}{\textbf{scene0712\_00}} & ATE (m) & \textbf{0.051} & 0.183 & 0.514 & 0.639 & 0.232 \\
 & RPE trans (m) & \textbf{0.016} & 0.021 & 0.025 & 0.017 & \textbf{0.016} \\
 & RPE rot (deg) & \textbf{0.383} & 3.807 & 1.943 & 2.221 & 0.619 \\
\midrule

\multirow{3}{*}{\textbf{scene0713\_00}} & ATE (m) & \textbf{0.204}  & 0.472 & 0.515 & 0.616 & 0.309 \\
 & RPE trans (m) & 0.124 & 0.090 & 0.047 & 0.047 & \textbf{0.024} \\
 & RPE rot (deg) & 9.536  & 18.216 & 3.437 & 4.127 & \textbf{1.287} \\
\midrule

\multirow{3}{*}{\textbf{scene0714\_00}} & ATE (m) & 0.891  & 0.644 & 0.389 & 0.916 & \textbf{0.372} \\
 & RPE trans (m) & 0.215 & 0.075 & 0.064 & 0.045 & \textbf{0.019} \\
 & RPE rot (deg) & 9.190 & 7.498 & 2.630 & 3.485 & \textbf{0.418} \\
\midrule

\multirow{3}{*}{\textbf{scene0715\_00}} & ATE (m) & 0.267  & \textbf{0.230} & 0.239 & 0.511 & 0.341 \\
 & RPE trans (m) & 0.156  & 0.039 & 0.049 & 0.039 & \textbf{0.026} \\
 & RPE rot (deg) & 15.059 & 8.835 & 2.930 & 3.524 & \textbf{0.611} \\
\midrule

\multirow{3}{*}{\textbf{scene0717\_00}} & ATE (m) & \textbf{0.091}  & 0.324 & 0.508 & 0.782 & 0.252 \\
 & RPE trans (m) & 0.040  & 0.050 & 0.058 & 0.040 & \textbf{0.022} \\
 & RPE rot (deg) & 0.586 & 7.267 & 3.006 & 3.453 & \textbf{0.555} \\
\midrule

\multirow{3}{*}{\textbf{scene0718\_00}} & ATE (m) & X & 0.350 & \textbf{0.111} & 0.385 & 0.295 \\
 & RPE trans (m) & X & 0.080 & 0.065 & 0.066 & \textbf{0.039} \\
 & RPE rot (deg) & X & 12.460 & 3.837 & 5.189 & \textbf{0.844} \\
\midrule

\multirow{3}{*}{\textbf{scene0719\_00}} & ATE (m) & \textbf{0.051}  & 0.373 & 0.171 & 0.657 & 0.268 \\
 & RPE trans (m) & 0.019 & 0.041 & 0.044 & 0.031 & \textbf{0.012} \\
 & RPE rot (deg) & \textbf{0.330} & 6.919 & 2.423 & 3.380 & 0.401 \\
\midrule

\multirow{3}{*}{\textbf{scene0720\_00}} & ATE (m) & \textbf{0.133}  & 0.390 & 0.331 & 0.389 & 0.815 \\
 & RPE trans (m) & \textbf{0.040}  & 0.034 & 0.030 & 0.027 & 0.021 \\
 & RPE rot (deg) & 0.900  & \textbf{0.668} & 2.002 & 1.915 & 0.875 \\
\midrule

\multirow{3}{*}{\textbf{scene0721\_00}} & ATE (m) & X  & 0.521 & \textbf{0.259} & 1.345 & 0.625 \\
 & RPE trans (m) & X  & 0.054 & \textbf{0.042} & 0.070 & 0.110 \\
 & RPE rot (deg) & X & 6.439 & \textbf{2.022} & 2.260 & 5.304 \\
\midrule

\multirow{3}{*}{\textbf{scene0722\_00}} & ATE (m) & \textbf{0.050}  & 0.427 & 0.319 & 0.486 & 0.467 \\
 & RPE trans (m) & 0.027 & 0.041 & 0.042 & 0.031 & \textbf{0.019} \\
 & RPE rot (deg) & \textbf{0.444} & 8.193 & 2.943 & 3.523 & 0.489 \\
\midrule

\multirow{3}{*}{\textbf{scene0723\_00}} & ATE (m) & \textbf{0.139}  & 0.766 & 0.483 & 0.521 & 0.220 \\
 & RPE trans (m) & 0.031  & 0.079 & 0.036 & 0.028 & \textbf{0.015} \\
 & RPE rot (deg) & 0.796 & 5.675 & 2.201 & 2.304 & \textbf{0.603} \\
\midrule

\multirow{3}{*}{\textbf{scene0724\_00}} & ATE (m) & \textbf{0.062}  & 0.647 & 0.429 & 0.702 & 0.332 \\
 & RPE trans (m) & 0.040 & 0.090 & 0.036 & 0.027 & \textbf{0.013} \\
 & RPE rot (deg) & \textbf{0.854} & 5.857 & 2.796 & 3.218 & 1.022 \\
\midrule

\multirow{3}{*}{\textbf{scene0725\_00}} & ATE (m) & X  & 0.714 & 0.550 & 0.882 & \textbf{0.548} \\
 & RPE trans (m) & X & 0.075 & 0.036 & 0.027 & \textbf{0.013} \\
 & RPE rot (deg) & X & 9.665 & 2.272 & 2.612 & \textbf{0.708} \\
\midrule

\multirow{3}{*}{\textbf{scene0726\_00}} & ATE (m) & \textbf{0.100}  & 0.474 & 0.258 & 0.380 & 0.193 \\
 & RPE trans (m) & 0.051 & 0.055 & 0.043 & 0.035 & \textbf{0.011} \\
 & RPE rot (deg) & 0.563 & 6.567 & 2.524 & 2.904 & \textbf{0.458} \\

\bottomrule

\end{tabular}}
\label{tab::scannet_perscene}
\end{center}
\vspace{-10pt}
\end{table}

\section{Additional Visualization}
\begin{figure*}[tb]
\centering
\begin{tabular}{cccc}
\centering
{\includegraphics[width=0.22\linewidth]{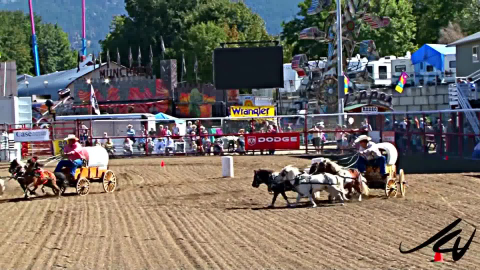}} & 
{\includegraphics[width=0.22\linewidth]{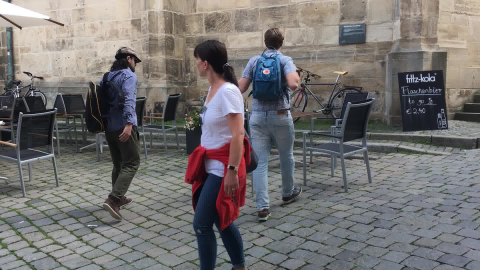}} &
{\includegraphics[width=0.22\linewidth]{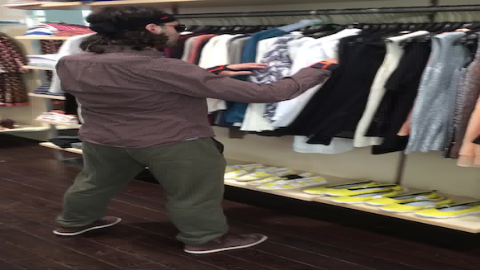}} &
{\includegraphics[width=0.22\linewidth]{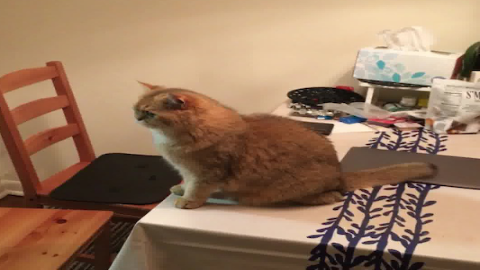}} \\

{\includegraphics[width=0.22\linewidth]{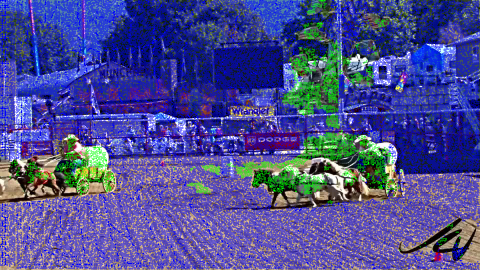}} & 
{\includegraphics[width=0.22\linewidth]{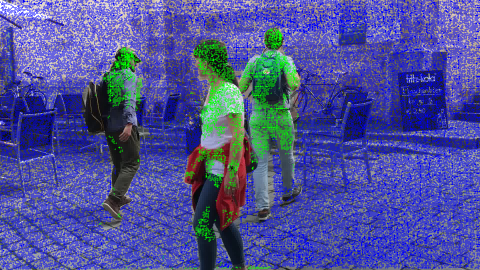}} & 
{\includegraphics[width=0.22\linewidth]{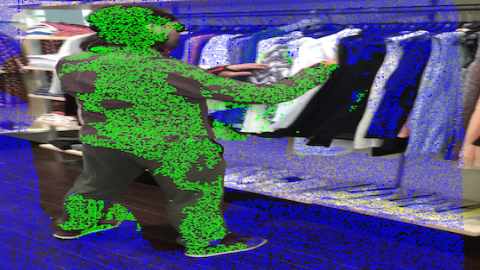}} &
{\includegraphics[width=0.22\linewidth]{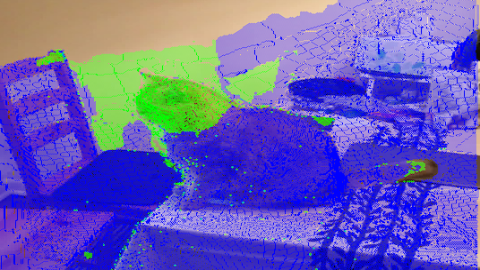}} \\

{\includegraphics[width=0.22\linewidth]{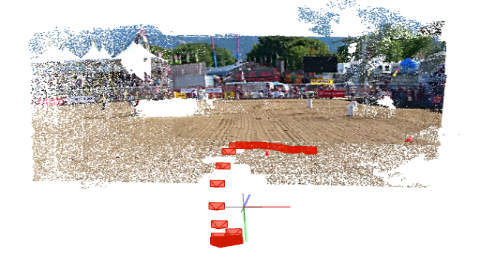}} & 
{\includegraphics[width=0.22\linewidth]{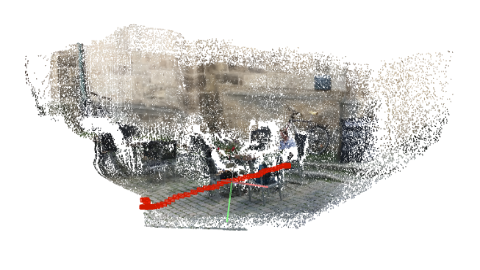}} & 
{\includegraphics[width=0.22\linewidth]{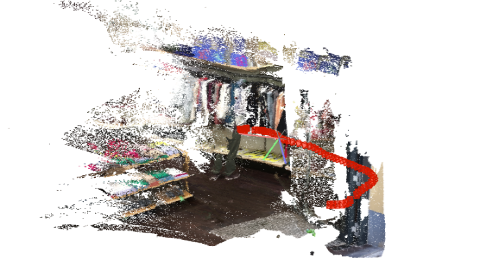}} &
{\includegraphics[width=0.22\linewidth]{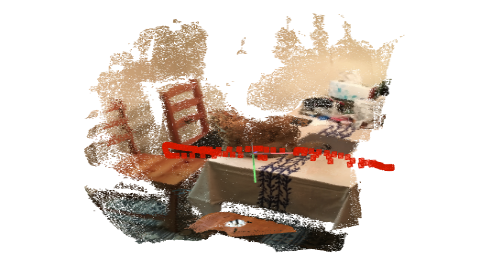}} \\

\\
\\
\\

{\includegraphics[width=0.22\linewidth]{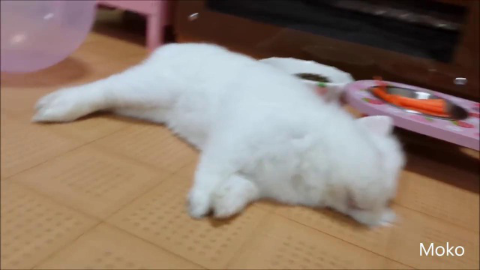}} & 
{\includegraphics[width=0.22\linewidth]{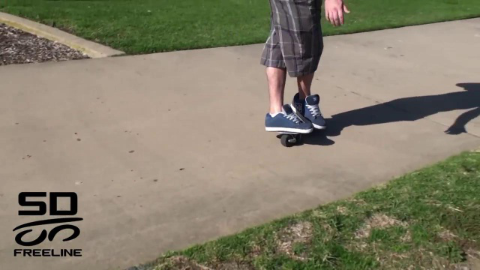}} &
{\includegraphics[width=0.22\linewidth]{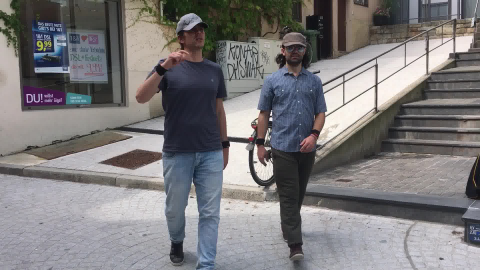}} &
{\includegraphics[width=0.22\linewidth]{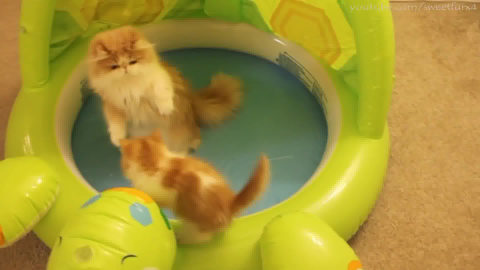}} \\

{\includegraphics[width=0.22\linewidth]{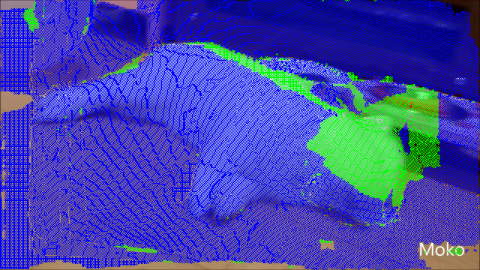}} & 
{\includegraphics[width=0.22\linewidth]{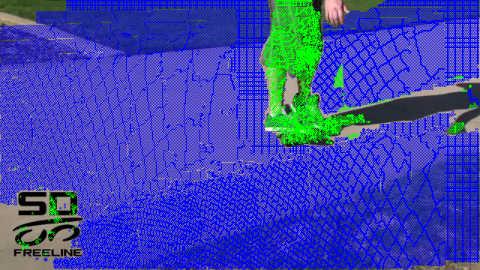}} & 
{\includegraphics[width=0.22\linewidth]{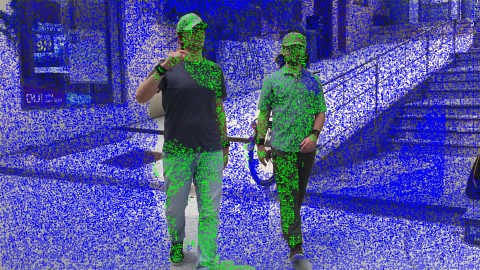}} &
{\includegraphics[width=0.22\linewidth]{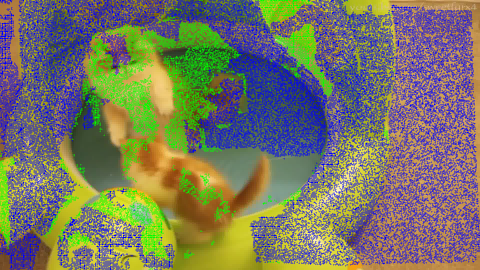}} \\

{\includegraphics[width=0.22\linewidth]{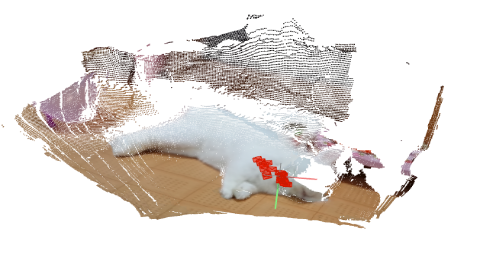}} & 
{\includegraphics[width=0.22\linewidth]{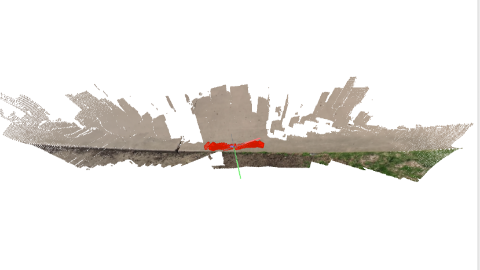}} & 
{\includegraphics[width=0.22\linewidth]{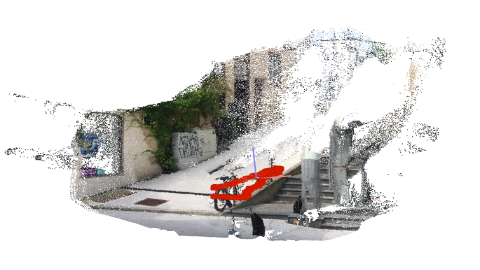}} &
{\includegraphics[width=0.22\linewidth]{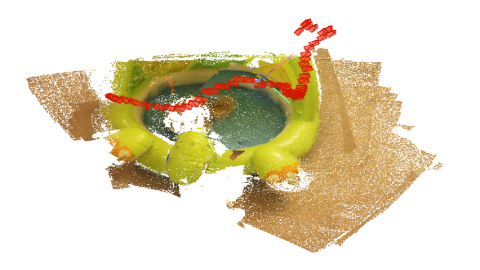}} \\

\\
\\
\\
{\includegraphics[width=0.22\linewidth]{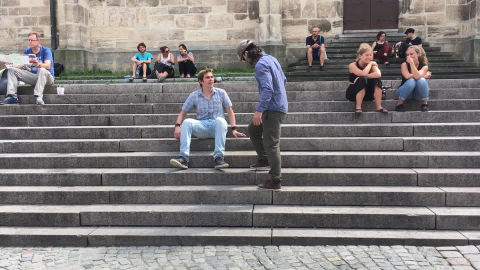}} & 
{\includegraphics[width=0.22\linewidth]{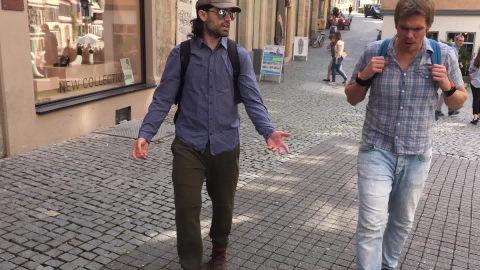}} &
{\includegraphics[width=0.22\linewidth]{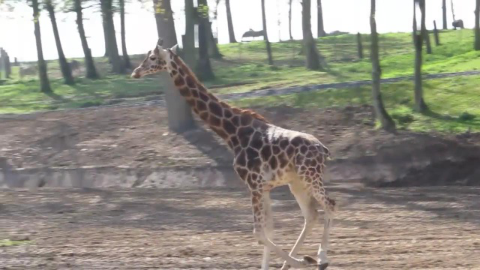}} &
{\includegraphics[width=0.22\linewidth]{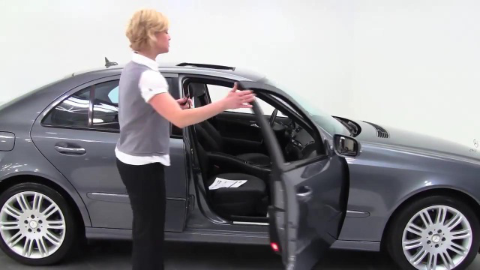}} \\

{\includegraphics[width=0.22\linewidth]{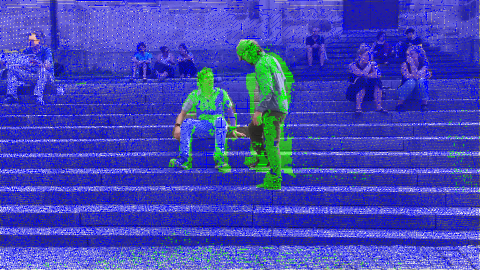}} & 
{\includegraphics[width=0.22\linewidth]{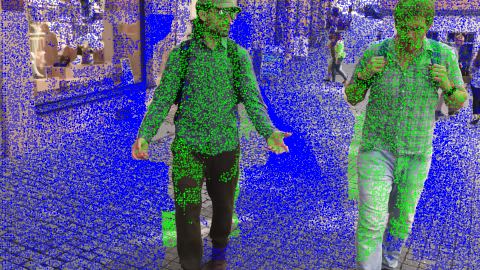}} & 
{\includegraphics[width=0.22\linewidth]{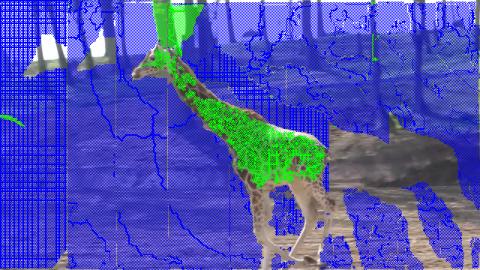}} &
{\includegraphics[width=0.22\linewidth]{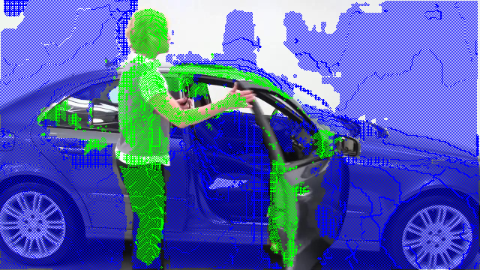}} \\

{\includegraphics[width=0.22\linewidth]{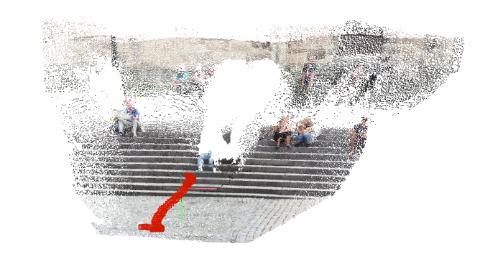}} & 
{\includegraphics[width=0.22\linewidth]{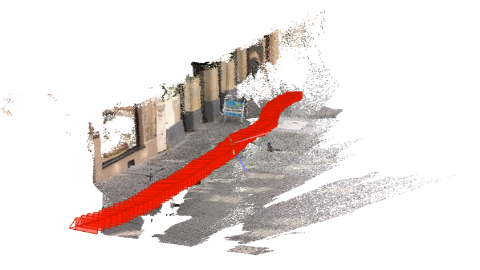}} & 
{\includegraphics[width=0.22\linewidth]{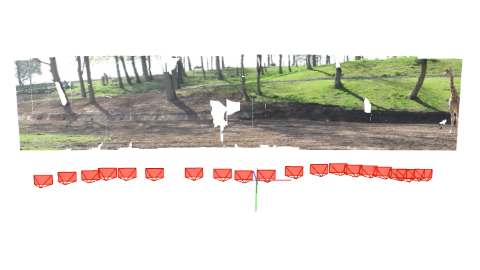}} &
{\includegraphics[width=0.22\linewidth]{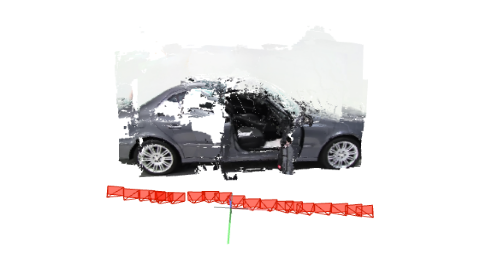}} \\

\end{tabular}

\caption{Visualization of trajectory motion segmentation and camera localization of in-the-wild videos. \textbf{Moving pixels from point trajectories are colored in green and static background pixels are in blue. Free space with no colored pixels indicates that there are no trajectory points due to occlusion or large flow forward-backward consistency error.}}
\label{fig::supp_wild}
\end{figure*}

We show more visualizations about trajectory motion segmentation and camera localization in Figure~\ref{fig::supp_wild}. Sequences are from 3DPW~\cite{von2018recovering}, Youtube-VOS~\cite{xu2018youtube}, GOT-10K~\cite{huang2019got} and BANMO~\cite{yang2021banmo} dataset. See the attached video for better experience.

\clearpage
%
%
\bibliographystyle{splncs04}
\bibliography{egbib}
\end{document}